\documentclass[conference]{IEEEtran}
\IEEEoverridecommandlockouts
\usepackage{cite}
\usepackage{amsmath,amssymb,amsfonts}
\usepackage{graphicx}
\usepackage{textcomp}
\usepackage{xcolor}

\usepackage{algorithm} 
\usepackage{algpseudocode}
\usepackage[switch]{lineno}

\usepackage{subcaption}
\usepackage{layouts}

\usepackage{hyperref}
\usepackage{url}

\newcommand{\enc}{f}
\newcommand{\z}{z}
\newcommand{\tr}{T}
\newcommand{\cont}{c}
\newcommand{\inv}{I}
\newcommand{\uncont}{u}
\newcommand{\obs}{s}
\newcommand{\obsspace}{\mathcal{S}}
\newcommand{\loss}{\mathcal{L}}

\usepackage{amsmath,amsfonts,bm}

\def\eqref#1{equation~\ref{#1}}

\def\1{\bm{1}}

\DeclareMathAlphabet{\mathsfit}{\encodingdefault}{\sfdefault}{m}{sl}
\SetMathAlphabet{\mathsfit}{bold}{\encodingdefault}{\sfdefault}{bx}{n}

\DeclareMathOperator*{\argmax}{arg\,max}

\DeclareMathOperator{\EX}{\mathbb{E}}

\def\BibTeX{{\rm B\kern-.05em{\sc i\kern-.025em b}\kern-.08em
    T\kern-.1667em\lower.7ex\hbox{E}\kern-.125emX}}
\begin{document}

\title{Disentangled (Un)Controllable Features\\
}

\author{\IEEEauthorblockN{1\textsuperscript{st} Jacob E. Kooi}
\IEEEauthorblockA{\textit{Quantitative Data Analytics} \\
\textit{Vrije Universiteit Amsterdam}\\
Amsterdam, Netherlands\\
j.e.kooi@vu.nl}
\and
\IEEEauthorblockN{2\textsuperscript{nd} Mark Hoogendoorn}
\IEEEauthorblockA{\textit{Quantitative Data Analytics} \\
\textit{Vrije Universiteit Amsterdam}\\
Amsterdam, Netherlands\\
m.hoogendoorn@vu.nl}
\and
\IEEEauthorblockN{3\textsuperscript{rd} Vincent Francois-Lavet}
\IEEEauthorblockA{\textit{Quantitative Data Analytics} \\
\textit{Vrije Universiteit Amsterdam}\\
Amsterdam, Netherlands\\
vincent.francoislavet@vu.nl}
}

\maketitle

\begin{abstract}
In the context of MDPs with high-dimensional states, downstream tasks are predominantly applied on a compressed, low-dimensional representation of the original input space. A variety of learning objectives have therefore been used to attain useful representations. However, these representations usually lack interpretability of the different features. We present a novel approach that is able to disentangle latent features into a controllable and an uncontrollable partition. We illustrate that the resulting partitioned representations are easily interpretable on three types of environments and show that, in a distribution of procedurally generated maze environments, it is feasible to interpretably employ a planning algorithm in the isolated controllable latent partition.
\end{abstract}

\section{Introduction}
Learning from high-dimensional data remains a challenging task. Particularly for reinforcement learning (RL), the complexity and high dimensionality of the Markov Decision Process (MDP) \cite{Bellman} states often leads to complex or intractable solutions.
In order to facilitate learning from high-dimensional input data, an encoder architecture can be used to compress the inputs into a lower-dimensional latent representation. To this extent, a plethora of work has successfully focused on discovering a compressed encoded representation that accommodates the underlying features for the task at hand \cite{Jonschkowski2015LearningPriors,Jaderberg2016ReinforcementTasks,Laskin2020CURLLearning,Lee2020PredictiveRL,Yarats2019ImprovingImages,Schwarzer2020Data-EfficientRepresentations,Kostrikov2020ImagePixels}.

\begin{figure*}[!t]

\vspace{-10mm}
\qquad \quad
     \begin{subfigure}[b]{0.45\textwidth}
         \centering

         \includegraphics[width=2in]{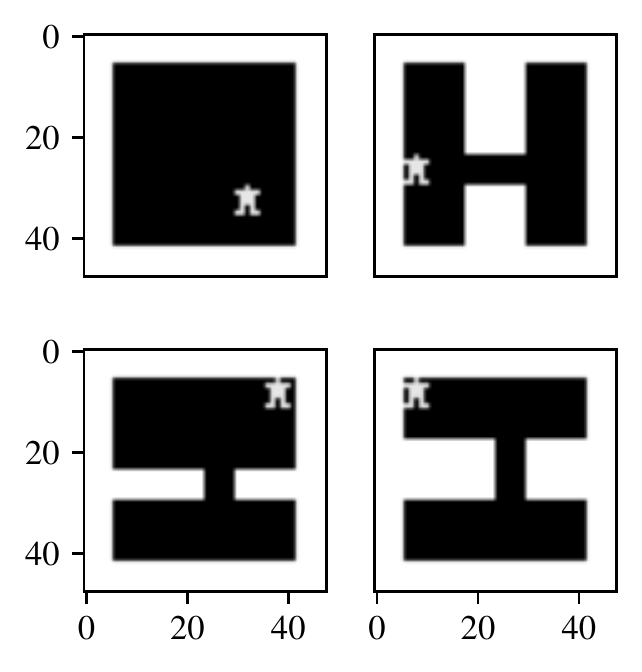}%
         
         \label{fig:y equals x}

     \end{subfigure}
     \begin{subfigure}[b]{0.45\textwidth}

         \includegraphics[width=2.7in]{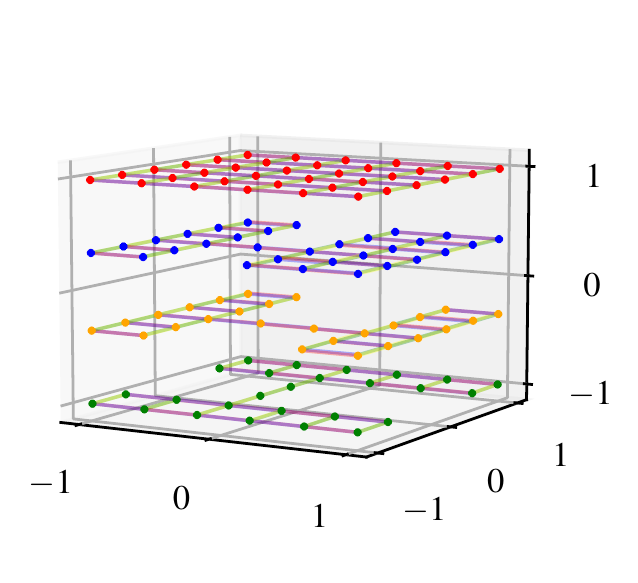}%

         \label{fig:three sin x}
     \end{subfigure}
        \caption{Visualization in a maze environment of four random pixel observations $s \in \mathbb{R}^{48\times48}$ (left) and the encoded observations $\z = \enc(\obs;\theta_{enc}) \hspace{2mm}  \forall s \in \mathcal{S}$ (right). On the right, we can see the disentanglement of the controllable latent $\z^{\cont} \in \mathbb{R}^{2}$ on the horizontal axes, and the uncontrollable latent $\z^{\uncont} \in \mathbb{R}^{1}$ on the vertical axis. The encoder is trained on high-dimensional tuples $(\obs_{t}, a_{t}, r_{t}, \obs_{t+1})$, sampled from a replay buffer $\mathcal{B}$, gathered from random trajectories in the four maze environments shown on the left. All possible states in all four mazes are encoded and plotted with the transition prediction for each possible action, revealing a clear disentanglement between the controllable latents (agent x-y position) and the uncontrollable latent (wall architecture). Note that all samples are taken from the same buffer, filled with samples from all four mazes.}
     \label{fig:fourmaze}
\end{figure*}

The resulting low-dimensional representations however seldom contain specific disentangled features, which leads to disorganized latent information. This means that the individual latent states can represent the information from the state in any arbitrary way. The result is a representation with poor interpretability, as the latent states cannot be connected to certain attributes of the original observation space (e.g, the x-y coordinates of the agent). Prior work in structuring a latent representation has shown notions and use of interpretability in MDP representations \cite{Francois-Lavet2019CombinedRepresentations}. When expanding this notion of interpretability to be compatible with RL, it has been argued that the controllable features should be an important element of a latent representation, since it generally represents what is directly influenced by the policy. In this light, \cite{Thomas2017IndependentlyFactors} have introduced the concept of isolating and disentangling controllable features in a low-dimensional maze environment, by means of a selectivity loss. Furthermore, \cite{Kipf2019ContrastiveModels} took an object-centric approach to isolate distinct objects in MDPs and \cite{weakly_supervised} showed theoretical foundations for this isolation in a weakly-supervised controllable setting. Controllable features however only represent a fragment of an environment, where in many cases the uncontrollable features are of equal importance. For example, in the context of a distribution of mazes, for the prediction of the next controllable (agent) state following an action, the information about the wall structure is crucial (see Fig.~\ref{fig:fourmaze}).
We therefore hypothesize that a thorough representation should incorporate controllable and uncontrollable features, ideally in a disentangled, interpretable arrangement; Intepretability is crucial for future real-world deployment \cite{Glanois_Survey}, while an additional benefit would be that the separation of the controllable and uncontrollable features can be exploited in downstream algorithms such as planning.

Our contribution consists of an algorithm that, showcased in three different MDP settings, explicitly disentangles the latent representation into a controllable and an uncontrollable latent partition. This is highlighted on three types of environments, each with a varying class of controllable and uncontrollable elements. This allows for a precise and visible separation of the latent features, improving interpretability, representation quality and possibly moving towards a basis for building causal relationships between an agent and its environment. The unsupervised learning algorithm consists of both an action-conditioned and a state-only forward predictor, along with a contrastive and an adversarial loss, which isolate and disentangle the controllable versus the non-controllable features. Furthermore, we show an application of learning and planning on the human-interpretable disentangled latent representation, where the properties of disentanglement allow the planning algorithm to operate solely in the controllable partition of the latent representation.

\section{Related Work}

\paragraph{General Representation Learning}
Many works have focused on converting high-dimensional inputs to a compact, abstract latent representation. Learning this representation can make use of auxiliary, unsupervised tasks in addition to the pure RL objectives \cite{Jaderberg2016ReinforcementTasks}. One way to ensure a meaningful latent space is to implement architectures that require a pixel reconstruction loss such as a variational \cite{Kingma2013Auto-EncodingBayes,Higgins2017DARLA:Learning} or a deterministic \cite{Yarats2019ImprovingImages} autoencoder. Other approaches combined reward reconstruction with latent prediction {\cite{Gelada2019DeepMDP:Learning}, pixel reconstruction with planning \cite{Hafner2019LearningPixels,Hafner2020MasteringModels} or used latent predictive losses without pixel reconstruction
\cite{Lee2020PredictiveRL,Schwarzer2020Data-EfficientRepresentations}.

\paragraph{Representing controllable features}

In representation learning for RL, a focus on controllable features can be beneficial as these features are strongly influenced by the policy \cite{Thomas2017IndependentlyFactors}. This can be done using generative methods \cite{BetaVAE_controllable_features}, but is most commonly pursued using an auxiliary inverse-prediction loss; predicting the action that was taken in the MDP \cite{Jonschkowski2015LearningPriors}. The work in \cite{Pathak2017Curiosity-drivenPrediction,Badia2020NeverStrategies} builds a latent representation with an emphasis on the controllable features of an environment with inverse-prediction losses, and uses these features to guide exploratory behavior. Furthermore,  \cite{Provable_RL_Exogenous} and concurrent work by \cite{multistep_inverse} employ multi-step inverse prediction to successfully encompass controllable features in their representation. However, these works have not expressed a focus on also retaining the uncontrollable features in their representation, which is a key aspect in our work.

\paragraph{Partitioning a latent representation}

Sharing similarity in terms of the separation of the latent representation, \cite{Cyclic_recon} disentangle the latent representation in the domain adaptation setting into a task-relevant and a context partition, by means of adversarial predictions with gradient reversals and cyclic reconstruction. \cite{Fu2021LearningAbstractions} use a reconstruction-based adversarial architecture that divides their latent representation into reward-relevant and irrelevant features. Related work by \cite{denoised_MDP} further divides the latent representation of Dreamer \cite{Hafner2019LearningPixels}, using action-conditioned and state-only forward predictors, into controllable, uncontrollable and their respective reward relevant and irrelevant features. As compared to \cite{denoised_MDP}, who focus on distraction-efficient RL, we purely focus on the representational learning aspect of these predictors, and show notions of separation in low-dimensional, structured representations of MDPs, leaning towards enhanced interpretability. Furthermore, we use an adversarial loss to enforce disentanglement between $\z^{\cont}$ and $\z^{\uncont}$, and apply a contrastive loss instead of pixel reconstruction to avoid representation collapse due to latent forward prediction.


\paragraph{Interpretable representations in MDPs}

More closely related to our research is the work by \cite{Thomas2017IndependentlyFactors}, which connects individual latent dimensions to independently controllable states in a maze using a reconstruction loss and a selectivity loss. The work by \cite{Francois-Lavet2019CombinedRepresentations} visualizes the representation of an agent and its transitions in a maze environment, but does not disentangle the agent state in its controllable and uncontrollable parts, which limits the interpretability analysis and does not allow simplifications during planning. The work by \cite{Kipf2019ContrastiveModels} uses an object-oriented approach to isolate different (controllable) features, using graph neural networks (GNN's) and a contrastive forward prediction loss, but does not discriminate between controllable and uncontrollable features. Further work in this direction by \cite{weakly_supervised} focuses on theoretical foundations for an encoder to structurally represent a distinct controllable object. We aim to progress the aforementioned lines of research by using a representation learning architecture that disentangles an MDP's latent representation into interpretable, disentangled controllable and uncontrollable features. Finally, we show that having separate partitions of controllable and uncontrollable features can be exploited in a planning algorithm. Exploitations like these are done in combination with prior knowledge of a certain MDP, as in \cite{group_symmetries}.

\begin{figure*}[!t]
\centering
\vspace{-14mm}

\resizebox{0.65\textwidth}{!}{
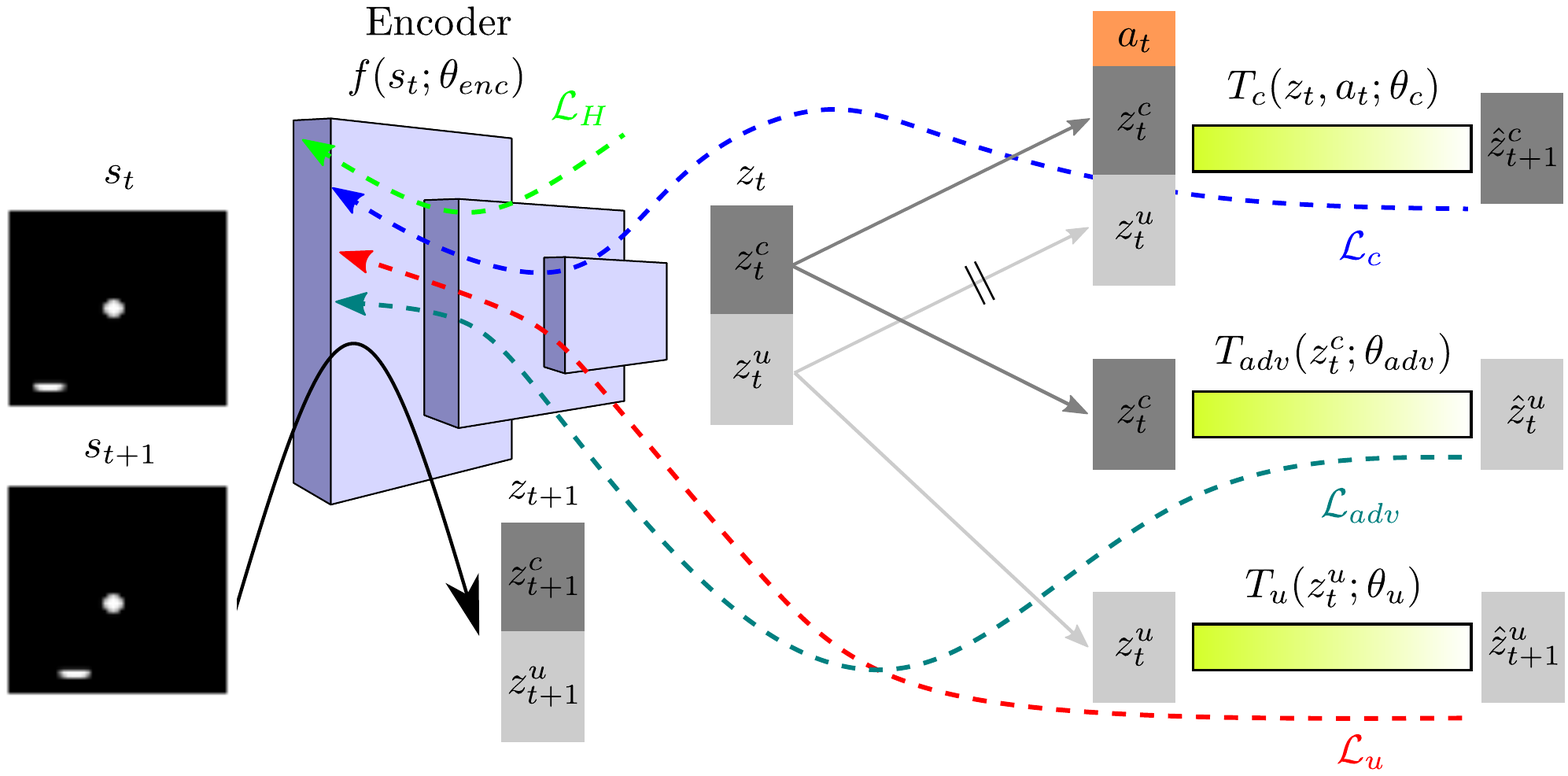
}
\hspace{4mm}

\vspace{-100mm}
\caption{Overview of the disentangling architecture, with dashed lines representing gradient propagation and green rectangles representing parameterized prediction functions. An observation $\obs_{t}$ is encoded into a latent representation consisting of two parts; $\z^{\cont}_{t}$ and $\z^{\uncont}_{t}$, which represent controllable and uncontrollable features respectively. These separated representations are then independently used to make action-conditioned, state-only and adversarial predictions in order to provide gradients to the encoder that disentangle the latent representation $\z_{t}$ into controllable ($\z^{\cont}_{t}$) and uncontrollable ($\z^{\uncont}_{t}$) partitions.}
\label{fig:catcher_adv_architecture}
\end{figure*}

\section{Preliminaries}
\label{prelims}

We consider an agent acting within an environment, where the environment is modeled as a discrete Markov Decision Process (MDP) defined as a tuple $(\mathcal{S}, \mathcal{A}, T, R, \gamma)$. Here, $\mathcal{S}$ is the state space, $\mathcal{A}$ is the action space, $T: \mathcal{S} \times \mathcal{A} \rightarrow \mathcal{S} $ is the environment's transition function, $R: \mathcal{S} \times \mathcal{A} \rightarrow \mathcal{R}$ is the environment's reward mapping and $\gamma$ is the discount factor. We consider the setting where we have access to a replay buffer ($\mathcal{B}$) of visited states $\obs_{t} \in \obsspace$ that were followed by actions $a_{t} \in \mathcal{A}$ and resulted in the rewards $r_{t} \in \mathcal{R}$ and the next states $\obs_{t+1}$. One entry in $B$ contains a tuple of past experience $(\obs_{t}, a_{t}, r_{t}, \obs_{t+1})$. The agent's goal is to learn a policy $\pi: \mathcal{S} \rightarrow \mathcal{A}$ that maximizes the expectation of the discounted return $V^\pi(s) = \EX _{\tau} [\sum_{t=0}^{\infty} \gamma^{t}R(s_{t},a_{t}) \mid s_t=s]$, where $\tau$ is a trajectory following the policy $\pi$.

Furthermore, we examine the setting where a high-dimensional state ($\obs_{t}\in \mathbb{R}^{v}$) is compressed into a lower-dimensional latent state $\z_{t}
\in \mathcal{Z} = \mathbb{R}^{w}$ where $\mathcal{Z}$ represents the latent space with $w \leq v$. This is done by means of a neural network encoding $f: \mathcal{S} \rightarrow \mathcal{Z}$ where $f$ represents the encoder.

\section{Algorithm}
We aim for an interpretable and disentangled representation of the controllable and uncontrollable latent features. We define controllable features as the characteristics of the MDP that are predominantly affected by any action $a \in \mathcal{A}$, such as the position of the agent in the context of a maze environment. The uncontrollable features are those attributes that are not or only marginally affected by the actions. We show that the proposed disentanglement is possible by designing losses and gradient propagation through two separate parts of the latent representation. Specifically, to assign controllable information to the controllable latent partition, the gradient from an action-conditioned forward predictor is propagated through it. To assign uncontrollable information to the uncontrollable latent partition, the gradient from a state-only forward predictor is propagated through it. The remaining details will be provided in the rest of this Section.


We consider environments with high-dimensional states, represented as pixel inputs. These pixel inputs are subsequently encoded into a latent representation $z_{t} = (\z^{\cont}, \z^{\uncont}) \in \mathcal{Z} \in \mathbb{R}^{n_{\cont}}+\mathbb{R}^{n_{\uncont}}$, with the superscripts $\cont$ and $\uncont$ representing the controllable and uncontrollable features, and the superscripts $n_{\cont}$ and $n_{\uncont}$ representing their respective dimensions. The compression into a latent representation $\mathcal{S} \rightarrow \mathcal{Z}$ is done by means of a convolutional encoder, parameterized by a set of learnable parameters ~$\theta_{enc}$~according to:
\begin{equation} \label{eq:encoder_output}
    \z_{t} = (\z^{\cont}_{t}, \z^{\uncont}_{t}) = \enc(\obs_{t};\theta_{enc}).
\end{equation}
\noindent An overview of the proposed algorithm is illustrated in Fig.~\ref{fig:catcher_adv_architecture} and the details are provided hereafter. In this section, all losses and transitions are given under the assumption of a continuous abstract representation and a deterministic transition function. The algorithm could be adapted by replacing the losses related to the internal transitions with generative approaches (in the context of continuous and stochastic transitions) or a log-likelihood loss (in the context of stochastic but discrete representations).

\subsection{Controllable Features}
To isolate controllable features in the latent representation, $\z^{\cont}_{t}$ is used to make an action-conditioned forward prediction in latent space. In the context of a continuous latent space and deterministic transitions, $\z^{\cont}$ is updated using a mean squared error (MSE) forward prediction loss 
$\loss_{c} =  \big|\hat{\z}^{\cont}_{t+1} - \z^{\cont}_{t+1}\big|^{2}$, where $\hat{\z}^{\cont}_{t+1}$ is the action-conditioned residual forward prediction of the parameterized function $\tr_{c}(z,a;\theta_{c}): \mathcal{Z} \times \mathcal{A} \rightarrow \mathcal{Z}$:
\begin{equation}  \label{eq:controllable_forward_predictor}
    \hat{\z}^{\cont}_{t+1} = \tr_{c}(\z_{t}, a_{t};\theta_{c}) + \z^{\cont}_{t}
\end{equation}
\noindent and the prediction target $z^{\cont}_{t+1}$ is part of the encoder output $\enc(\obs_{t+1};\theta_{enc})$. Note that the full latent state $\z_{t}$ is necessary in order to predict $\hat{\z}^{\cont}_{t+1}$ (e.g. the uncontrollable features could represent a wall or other static structure that is necessary for the prediction of the controllable features). Furthermore, the uncontrollable latent partition input $\z^{\uncont}_{t}$ is accompanied by a stop gradient to discourage the presence of controllable features in $\z^{\uncont}$. 
When minimizing $\loss_{c}$, both the encoder ($\theta_{enc}$) as well as the predictor ($\theta_{c}$) are updated, which allows shaping the representation $\z^{\cont}$ as well as learning the internal dynamics.

\subsection{Uncontrollable Features}

To express uncontrollable features in the latent space, $\z^{\uncont}_{t}$ is used to make a state-only (not conditioned on the action $a_{t}$) forward prediction in latent space. This enforces uncontrollable features within the uncontrollable latent partition $\z^{\uncont}$, since features that are action-dependent cannot be accurately predicted with the preceding state only. Following a residual prediction, $\z^{\uncont}$ is then updated using a MSE forward prediction loss $\loss_{u} =  \big|\hat{\z}^{\uncont}_{t+1} - \z^{\uncont}_{t+1}\big|^{2}$, with $\hat{\z}^{\uncont}_{t+1}$ defined as:
\begin{equation}  \label{eq:uncontrollable_forward_predictor}
    \hat{\z}^{\uncont}_{t+1} = \tr_{u}(\z^{\uncont}_{t};\theta_{u}) + \z^{\uncont}_{t}
\end{equation}
\noindent and $\tr_{u}(\z^{u};\theta_{u}): \mathcal{Z} \rightarrow \mathcal{Z}$ representing the parameterized prediction function.~The target $\z^{\uncont}_{t+1}$ is part of the output of the encoder $ \enc(\obs_{t+1};\theta_{enc})$. 
When minimizing $\loss_{u}$, both $\theta_{enc}$ and $\theta_{u}$ are updated. In this way the loss $\loss_{u}$ drives the latent representation $\z^{\uncont}$, which is conditioned on $\theta_{enc}$ according to $(\z^{\cont}_{t}, \z^{\uncont}_{t}) = \enc(\obs_{t};\theta_{enc})$, to only represent the features of $s_{t}$ that are not conditioned on the action $a_{t}$.

\subsection{Avoiding Predictive Representation Collapse}\label{entropy}

Minimizing a forward prediction loss in latent space $\mathcal{Z}$ is prone to collapse \cite{Francois-Lavet2019CombinedRepresentations,Gelada2019DeepMDP:Learning}, due to the convergence of $\loss_{c}$ and $\loss_{u}$ when $\enc(\obs_{t};\theta_{enc})$ is a constant \hspace{1.5mm} $\forall \hspace{2mm} \obs_{t} \in \obsspace$.  
To avoid representation collapse when using forward predictors, a contrastive loss is used to enforce sufficient diversity in the latent representation:
 \begin{equation}  \label{eq:encoder_entropy}
    \loss_{H_1} = exp\big(-C_{d}\big\|\z_{t} - \bar{\z}_{t}\big\|_{2}\big)
\end{equation}
\noindent where $C_{d}$ represents a constant hyperparameter and $\bar{\z}_{t}$ is a `negative' batch of latent states $\z_{t}$, which is obtained by shifting each position of latent states in the batch by a random number between 0 and the batch size. In the random maze environment, an additional contrastive loss is added to further diversify the controllable representation:
 \begin{equation}  \label{eq:encoder_entropy_individual}
    \loss_{H_2} = exp\big(-C_{d}\big\|\z^{\cont}_{t} - \bar{\z}^{\cont}_{t}\big\|_{2}\big)
    \end{equation}
\noindent where $\z^{\cont}_{t}$ is obtained from randomly sampled trajectories. This additional regularizer proved neccessary to avoid collapse of $\z^{\cont}$ when moving to a near infinite number of possible mazes. More information on this subject can be found in Appendix~\ref{AppendixA4_entropy_randommaze}. The resulting contrastive loss $\loss_{H}$ for the random maze environment then consists of $0.5 \loss_{H_{1}} + 0.5 \loss_{H_{2}}$. The total loss used to update the encoder's parameters now consists of $\loss_{enc} = \loss_{c} + \loss_{u} + \loss_{H}$.

 \subsection{Guiding Feature Disentanglement with Adversarial Loss} \label{adversarial}   
 
When using a controllable latent space $\z^{\cont} \in \mathbb{R}^{x}, x \in \mathbb{N}$, where $x >g$, with $g$ representing the number of dimensions needed to portray the controllable features, some information about the uncontrollable features in the controllable latent representation might be present (see Appendix~\ref{appendix:catcher}). This is due to the non-enforcing nature of $\loss_{c}$, as the uncontrollable features are equally predictable with or without the action. 
To ensure that no information about the uncontrollable features is kept in the controllable latent representation, an adversarial component is added to the architecture in Fig.~\ref{fig:catcher_adv_architecture}. This is done by updating the encoder with an adversarial loss $\loss_{adv}$ and reversing the gradient \cite{Ganin2016Domain-AdversarialNetworks}. The adversarial loss is defined as
\begin{equation}  \label{eq:adversarial_loss}
    \loss_{adv} =  \big|\hat{\z}^{\uncont}_{t} - \z^{\uncont}_{t}\big|^{2},
\end{equation}
\noindent with $\hat{\z}^{\uncont}_{t} = \tr_{adv}(\z^{\cont}_{t};\theta_{adv})$, where $\hat{\z}^{\uncont}_{t}$ is the uncontrollable prediction of the parameterized function $\tr_{adv}(z^{\cont};\theta_{adv}) :\mathcal{Z} \rightarrow \mathcal{Z}$ and $\z^{\uncont}_{t}$ is the target.~Intuitively, since the parameters of  $\tr_{adv}(\z^{\cont} ;\theta_{adv})$ are being updated with $\loss_{adv}$ and the parameters of $\enc(s;\theta_{enc})$ are being updated with $-\loss_{adv}$, the prediction function can be seen as the discriminator and the encoder can be seen as the generator \cite{Goodfellow2014GenerativeNetworks}. The discriminator tries to give an accurate prediction of the uncontrollable latent $\z^{\uncont}$ given the controllable latent $\z^{\cont}$, while the generator tries to counteract the discriminator by removing any uncontrollable features from the controllable representation. In our case, the predictor is a multi-layer perceptron (MLP), which means that minimizing $\loss_{adv}$ enforces that no nonlinear relation between $\z^{\cont}$ and $\z^{\uncont}$ can be learned. We hypothesize that this is a deterministic approximation of minimizing the Mutual Information (MI) between $z^{u}$ and $z^{c}$. When using the adversarial loss, the combined loss propagating through the encoder consists of $\loss_{enc} = \loss_{c} + \loss_{u} + \loss_{H} - \loss_{adv}$. Here the minus term in $-\loss_{adv}$ represents a gradient reversal to the encoder. Note that the losses are not scaled, as this did not prove to be necessary for the experiments conducted.

\begin{algorithm}
	\caption{Disentangled (Un)Controllable Features} 
	\begin{algorithmic}[1]
        \State Initialize $\theta_{enc}$, $\theta_{c}$, $\theta_{u}$, $\theta_{adv}$
		\For {$iteration=1,2,\ldots, N$}
			\State Sample batch of tuples \{$s_{t}, a_{t}, s_{t+1}$\}
			\State Encode observations:  $\enc(s;\theta_{enc}) = \{z^{c}, z^{u}\}$
			\State Predict $\hat{z}^{c}_{t+1} = T_{c}(z^{c}_{t}, z^{u}_{t}, a; \theta_{c}) + z^{c}_{t} \hspace{20mm}$ \hspace{-15mm}// detach $z^{u}_{t}$
            \State Predict $\hat{z}^{u}_{t+1} = T_{u}( z^{u}_{t}; \theta_{u}) + z^{u}_{t} $
            \State Predict $\hat{z}^{u}_{t} = T_{adv}( z^{c}_{t}; \theta_{adv})$
            \State Compute losses $\mathcal{L}_{c}, \mathcal{L}_{u}, -\mathcal{L}_{adv}, \mathcal{L}_{H}$ \hspace{6mm}
            \State Update parameters $\theta_{enc}$, $\theta_{c}$, $\theta_{u}$, $\theta_{adv}$
		\EndFor	
    \end{algorithmic} \end{algorithm}
    \label{algo}

 \subsection{Downstream Tasks}

By disentangling a latent representation in a controllable and an uncontrollable part, one can more readily obtain human-interpretable features.~While interpretability is generally an important aspect, it is also important to test how a notion of human interpretability affects downstream performance, as it is generally desired to strike a good balance between interpretability and performance.
This is examined by training an RL agent on the learned and subsequently frozen latent representation. The action $a_{t}$ is chosen following an $\epsilon$-greedy policy, where a random action is taken with a probability $ \epsilon$, and with $(1-\epsilon)$ probability the policy $\pi(z) = \underset{a \in \mathcal{A}}{\argmax} \hspace{1mm} Q(\z, a;\theta)$ is evaluated, where $Q(\z, a;\theta)$ is the Q-network trained by Deep Double Q-Learning (DDQN) \cite{mnih2015humanlevel,vanHasselt2015DeepQ-learning}. The Q-network is trained with respect to a target $Y_{t}$:
\begin{equation}
    Y_t = r_{t} + \gamma Q(\z_{t+1}, \argmax_{a \in \mathcal{A}} Q(\z_{t+1}, a;\theta);\theta^{-}) \,.
\end{equation}
With $\gamma$ representing the environment's discount factor and $\theta^{-}$ the target Q-network's parameters. The target Q-network's parameters are updated as an exponential moving average of the original parameters $\theta$ according to: $\theta^{-}_{k+1} = (1-\tau)\theta^{-}_{k} + \tau \theta_{k}$,
where subscript $k$ represents a training iteration and $\tau$ represents a hyperparameter controlling the speed of the parameter update. The resulting DDQN loss is defined as $\loss_{Q} =  \big|Y_{t} - Q(\z_{t}, a; \theta)\big|^{2}$. The full computation of all losses is shown in pseudocode in \hyperref[algo]{Algorithm 1}.

\section{Experiments}
In this section, we showcase the disentanglement of controllable and uncontrollable features on three different environments, the complexity of which is in line with prior work on structured representations \cite{Thomas2017IndependentlyFactors,Disentangled,Francois-Lavet2019CombinedRepresentations,Kipf2019ContrastiveModels,weakly_supervised}: (i) a quadruple maze environment, (ii) the catcher environment and (iii) a randomly generated maze environment. The first environment yields a state space of 119 different observations, 
and is used to showcase the algorithm's ability to disentangle a low-dimensional latent representation. The catcher environment examines a setting where the uncontrollable features are not static, and the random maze environment is used to showcase disentanglement in a more complex distribution of over 25 million possible environments, followed by the application of downstream tasks by applying reinforcement learning (DDQN) and a latent planning algorithm running in the controllable latent partition . 
The base of the encoder is derived from \cite{Tassa2018DeepMindSuite} and consists of two convolutional layers, followed by a fully connected layer for low-dimensional latent representations or an
additional CNN for a higher-dimensional latent representation such as a feature map. For the full network architectures, we refer the reader to Appendix~\ref{app:architecture}.
In all environments, the encoder $\enc(s;\theta_{enc})$ is trained from a buffer $\mathcal{B}$ filled with transition tuples $(\obs_{t}, a_{t}, r_{t}, \obs_{t+1})$ from random trajectories. Note that, in interpretability, there is generally not a specific metric to optimize for. In order to produce interpretable representations, finding the right hyperparameters required manual (human) inspection of the plotted latent representations. An ablation of the hyperparameters used can be found in Appendices \hyperref[Appendix]{A1-A3}

\subsection{Quadruple Maze Environment}
\label{sec:maze_env1}
The maze environment consists of an agent and a selection of four distinct, handpicked wall architectures. The environment's state is provided as pixel observations $\obs_{t} \in \mathbb{R}^{48 \times 48}$, where an action moves the agent by 6 pixels in each direction (up, down, left, right) except if this direction is obstructed by a wall. We consider the context where there is no reward ($r_{t} = 0 \hspace{2mm} \forall \hspace{2mm} (\obs_{t}, a_{t}) \in (\mathcal{S}, \mathcal{A})$) and there is no terminal state.

We select a two-dimensional controllable representation ($\z^{\cont} \in \mathbb{R}^{2}$) and a one-dimensional uncontrollable representation ($\z^{\uncont} \in \mathbb{R}^{1}$).
The remaining hyperparameters and details can be found in Appendix~\ref{Appendix:hyperparams}.
The experiments are conducted using a buffer $\mathcal{B}$ filled with random trajectories from the four different basic maze architectures. The encoder's parameters are updated using $\loss_{enc}$ in Section~\ref{entropy} with $\loss_{H} = \loss_{H_{1}}$. After 50k training iterations, a clear disentanglement between the controllable ($\z^{\cont}$) and uncontrollable ($\z^{\uncont}$) latent representation can be seen in Fig.~\ref{fig:fourmaze}. One can observe that the encoder is updated so that the one-dimensional latent representation $\z^{\uncont}$ learns different values that define the type of wall architecture. A progression to this representation is provided in Appendix~\ref{appendix:fourmaze}.

\begin{figure}[!t]
\vspace{-7mm}
\hspace{6mm}
  \begin{subfigure}[b]{0.36\columnwidth}
    \subcaptionbox{Without $\loss_{adv}$}{
    \includegraphics[width=\textwidth]{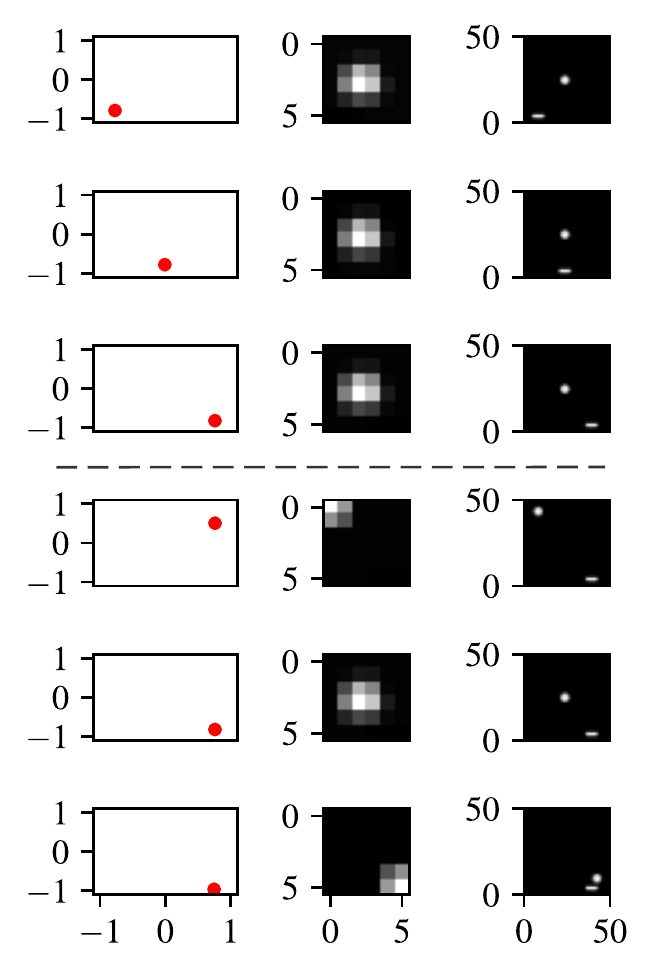}\hspace*{2em}%
    }
  \end{subfigure}
  \hspace{5mm}
  \begin{subfigure}[b]{0.36\columnwidth}        \subcaptionbox{With $\loss_{adv}$}{
    \includegraphics[width=\textwidth]{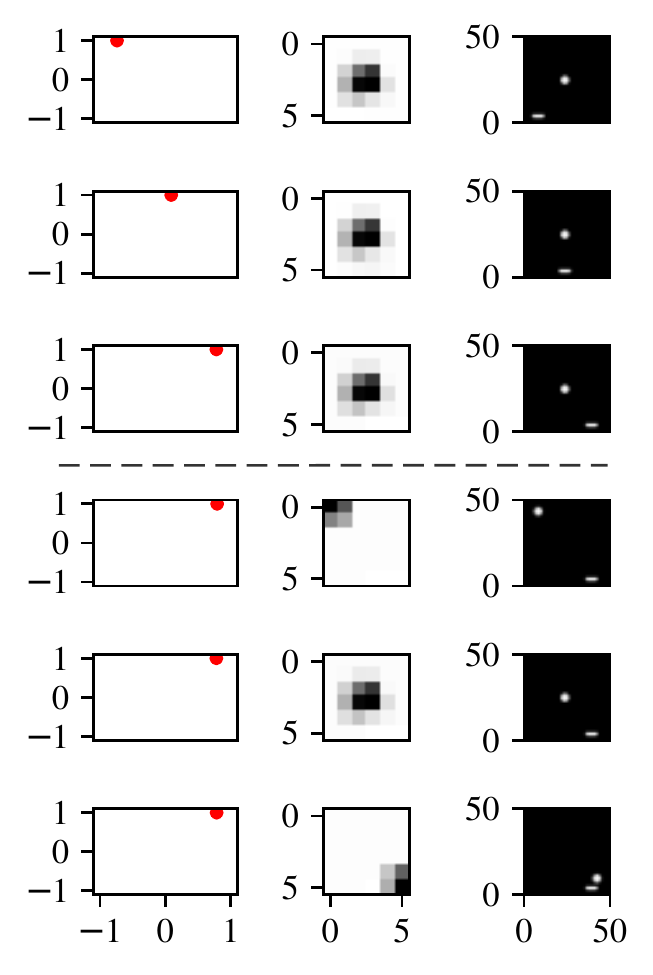}\hspace*{1em}%
    }
  \end{subfigure}
      \caption{Visualization of the latent feature disentanglement in the catcher environment after 200k training iterations, with $\z_{t} = \enc(\obs_{t};\theta_{enc})$ $\in \mathbb{R}^{2} +\mathbb{R}^{6\times 6}$. In (a) and (b), the left column shows $\z^{\cont}_{t}$, the middle column is a feature map representing $\z^{\uncont}_{t}$ and the right column is the pixel state $\obs_{t}$. The dashed lines separate observations where the ball position or the paddle position is kept fixed for illustration purposes. $\z^{\cont}$ tracks the agent position while $\z^{\uncont}$ tracks the falling ball. In b), note that even when having a two-dimensional controllable state (only 1 is needed, see Appendix~\ref{appendix:catcher}), the adversarial loss in b) makes sure that distinct ball positions have a negligible effect on $\z^{\cont}$ (left column), even when the high-level features of the agent and the ball might be hard to distinguish.}
      \label{fig:catcher2state}
      \end{figure}

      \subsection{Catcher Environment}
As opposed to the maze environment, the catcher environment encompasses uncontrollable features that are non-stationary. The ball is dropped randomly at the top of the environment and is falling irrespective of the actions, while the paddle position is directly modified by the actions. The environment's states are defined as pixel observations $\obs_{t}$ of size $\mathbb{R}^{51\times 51}$. At each time step, the paddle moves left or right by 3 pixels. Since we are only doing unsupervised learning, we consider the context where there is no reward ($r_{t} = 0 \hspace{2mm} \forall \hspace{2mm} (\obs_{t}, a_{t}) \in (\mathcal{S}, \mathcal{A})$) and an episode ends whenever the ball reaches the paddle or the bottom.

We take $\z^{\cont} \in \mathbb{R}^{2}$ and $\z^{\uncont} \in \mathbb{R}^{6\times 6}$. To test disentanglement, $\z^{\cont}$ is of a higher dimension than needed since the paddle (agent) only moves on the x-axis and would therefore require only one feature (see Appendix~\ref{appendix:catcher} for the simpler setting with $\z^{\cont} \in \mathbb{R}^{1}$). To show disentanglement, the redundant dimension of $\z^{\cont}$ should not or negligibly have information about $\z^{\uncont}$. The encoder's parameters are updated using $\loss_{enc}$ in Section~\ref{adversarial} with $\loss_{H} = \loss_{H_{1}}$. After training the encoder for 200k iterations, a selection of state observations $\obs_{t}$ and their encoding into the latent representation $\z = (\z^{\cont}, \z^{\uncont})$  can be seen in Fig.~\ref{fig:catcher2state}. A clear distinction between the ball and paddle representations can be observed, with the former residing in $\z^{\uncont}$ and the latter in $\z^{\cont}$.  

\subsection{Random Maze Environment}
\label{sec:maze_env2}
The random maze environment is similar to the maze environment from Section~\ref{sec:maze_env1}, but consists of a large distribution of randomly generated mazes with complex wall structures. The environment's state is provided as pixel observations $\obs_{t} \in \mathbb{R}^{48 \times 48}$, where an action moves the agent by 6 pixels in each direction. We consider $\z^{\cont} \in \mathbb{R}^{2}$ and $\z^{\uncont} \in \mathbb{R}^{6 \times 6}$. This environment tests the generalization properties of a disentangled latent representation, as there are over $25$ million possible maze architectures, corresponding to a probability of less than $4\cdot 10^{-8}$ to sample the same maze twice.
Note that because $\z^{\cont}$ is 2-dimensional, results with and without adversarial loss are in practice extremely close. After 50k training iterations, the latent representation $\z = (\z^{\cont}, \z^{\uncont})$ shows an interpretable disentanglement between the controllable and the uncontrollable features (see Fig.~\ref{fig:final_reps}). A clear distinction between the agent and the wall structure can be found inside $\z^{\cont}$ and $\z^{\uncont}$. Note that Instead of using a single dimension to `describe' the uncontrollable features $\z^{\uncont}$ (see Fig.~\ref{fig:fourmaze}), using a feature map for $\z^{\uncont}$ allows training an encoding that provides a more interpretable representation of the actual wall architecture.  

\begin{figure} [!t]
\vspace{-5mm}
     \centering
          \begin{subfigure}[b]{0.31\textwidth}
         \centering
         \includegraphics[width=\textwidth]{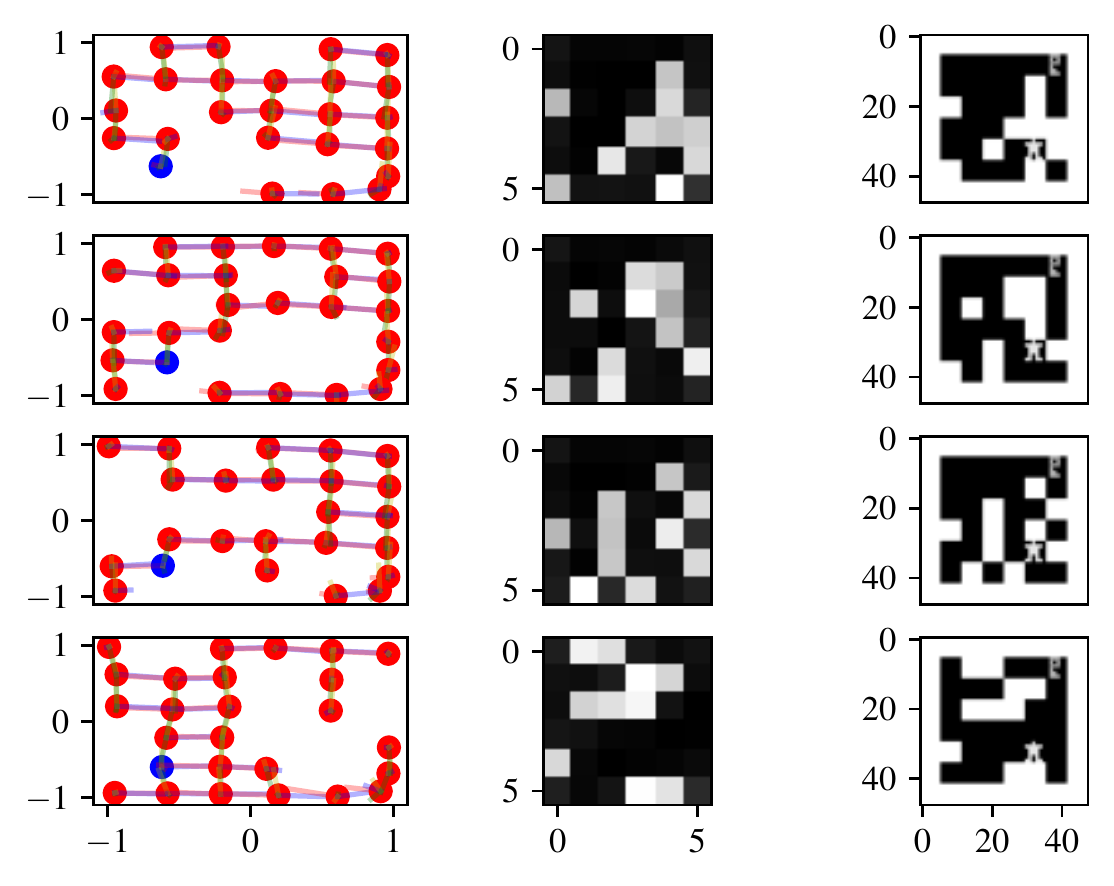}
         \caption{$\loss_{c} = \loss_{c}$}
         \label{fig:final_reps}
     \end{subfigure}
     \hfill
     \begin{subfigure}[b]{0.31\textwidth}
         \centering
         \includegraphics[width=\textwidth]{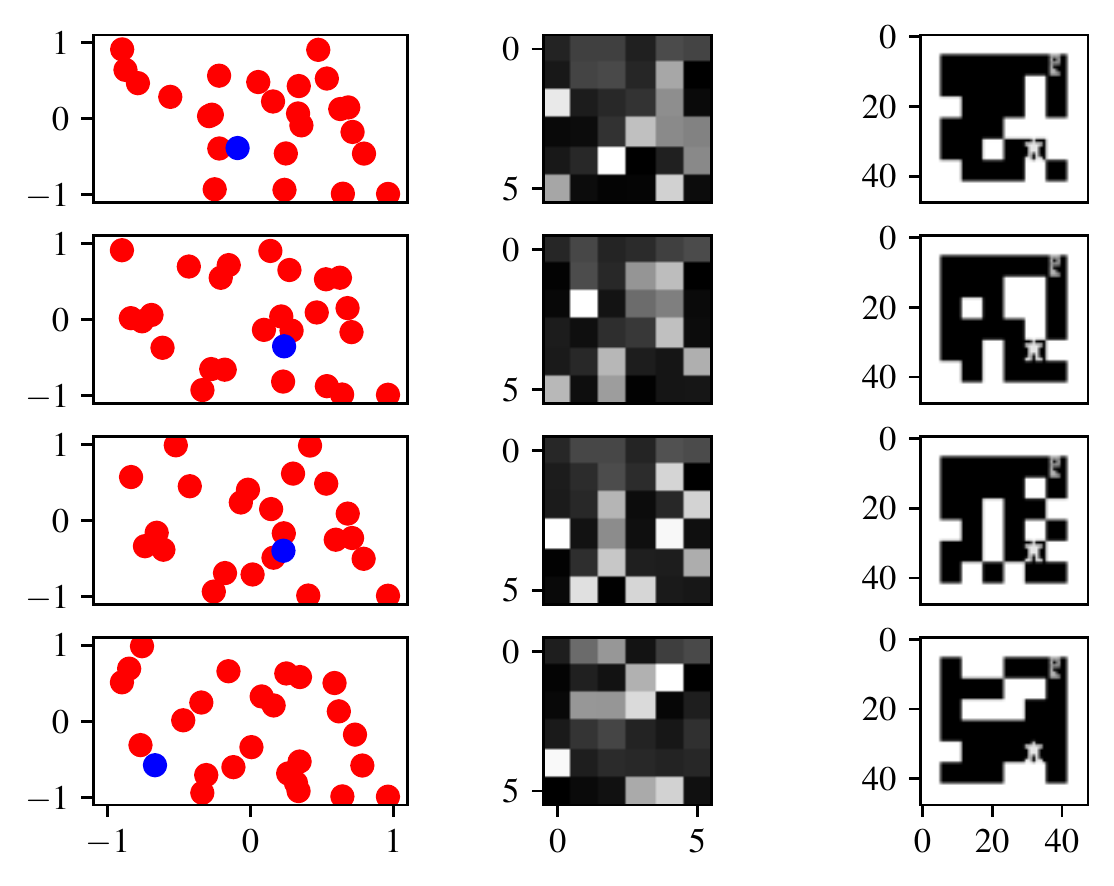}
         \caption{$\loss_{c} = \loss_{inv}$}
         \label{fig:inverse_reps}
     \end{subfigure}
     \hfill
     \begin{subfigure}[b]{0.31\textwidth}
         \centering
         \includegraphics[width=\textwidth]{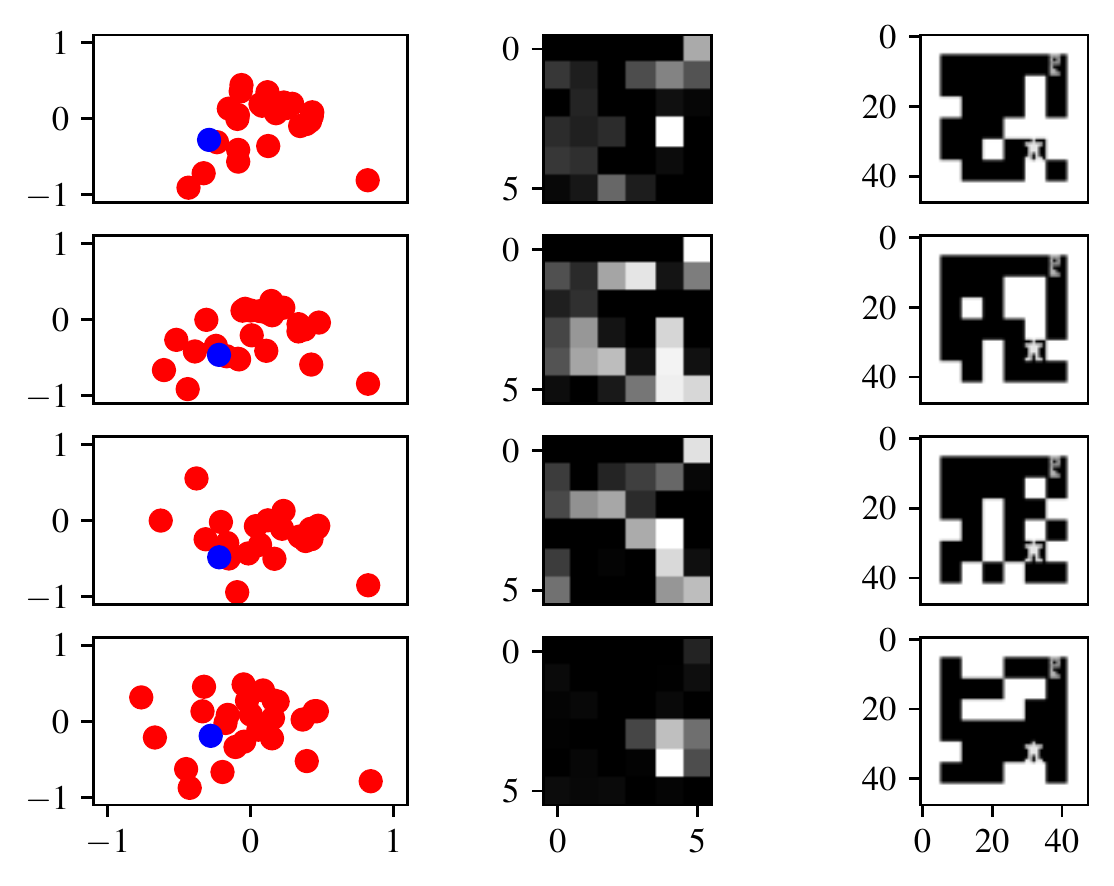}
         \caption{$\loss_{enc} = \loss_{Q}$}
         \label{fig:DDQN_reps}
     \end{subfigure}
     \caption{A plot of the latent representation for all observations in a single randomly sampled maze when training with the aforementioned losses (a), substituting the action-conditioned forward-prediction loss $\loss_{\cont}$ for an inverse-prediction loss $\loss_{inv}$ (b) and when end-to-end updating the encoder with only the Q-loss $\loss_{Q}$ from DDQN for 500k iterations (c). The left column shows the controllable latent $\z^{\cont}_{t} \in \mathbb{R}^{2}$ with the current state in blue, the remaining states in red, and the predicted movement due to actions as different colored bars for each individual action. The middle column shows the uncontrollable latent $\z^{\uncont}_{t} \in \mathbb{R}^{6 \times 6}$ and the right column shows the original state $s_{t} \in \mathbb{R}^{48 \times 48}$. Evidently, the controllable representations in (b) and (c) lack disentanglement and interpretability. Furthermore, the representation in (c) seems to have very little structure at all, showing that a representation that is optimized without prior structural incentives will often represent a black box.}
      \label{fig:inverse_DDQN_reps}
\end{figure}

\paragraph{Using an Inverse Predictor}
An alternative to the state-action forward prediction method used throughout the paper is the inverse (action) prediction loss.
An inverse prediction loss is often referred to in previous work that focuses on controllable features \cite{Jonschkowski2015LearningPriors,Pathak2017Curiosity-drivenPrediction,Badia2020NeverStrategies}. A single-step inverse prediction loss is defined as:
\begin{equation}
    \hat{a}_{t} = \inv(\z^{\cont}_{t}, \z^{\cont}_{t+1}, \z^{\uncont}_{t}; \theta_{inv}).
\end{equation} \label{eq:inverse}

\noindent Here, $\hat{a}_{t}$ is the predicted action and $\inv(\z^{\cont}_{t}, \z^{\cont}_{t+1}, \z^{\uncont}_{t}; \theta_{inv}): \mathcal{Z} \rightarrow 
\mathcal{A}$ is the inverse prediction network. To see whether an inverse predictor can generate structured, controllable representations in the random maze environment, we replace the action-conditioned forward predictor with an inverse predictor, so that $\z^{\cont}$ is no longer updated with $\loss_{c}$ but with $\loss_{inv}$ (see Appendix~\ref{appendix:Inverse} for details on $\loss_{inv}$).

The resulting representation can be seen in Fig.~\ref{fig:inverse_reps}. It seems that using $\loss_{inv}$, causes an absence of interpretable structure in the controllable latent representation $\z ^{\cont}_{t}$. Furthermore, there is a less precise disentanglement between the controllable and uncontrollable features, as differences can be observed in $\z ^{\cont}_{t}$ when encoding equal agent positions as pixel states $\obs_{t}$. In addition, an inverse predictor does not allow forward prediction in latent space, which can be used for planning as shown hereafter. It thus seems that in some environments, an inverse prediction loss might be insufficient to isolate the controllable features. Take for example the maze agent in the top-right maze of Fig.~\ref{fig:inverse_DDQN_reps}, where the agent can only move in the left direction. Even when using the wall information ($\z^{\uncont}_{t})$, an inverse predictor will not be able to predict the action taken when the agent does not go left. However, an action-conditioned forward predictor is able to predict the next state correctly regardless of which action was taken.

\begin{figure}[!t]
\hspace{-25mm}

\includegraphics[width=3.2in]{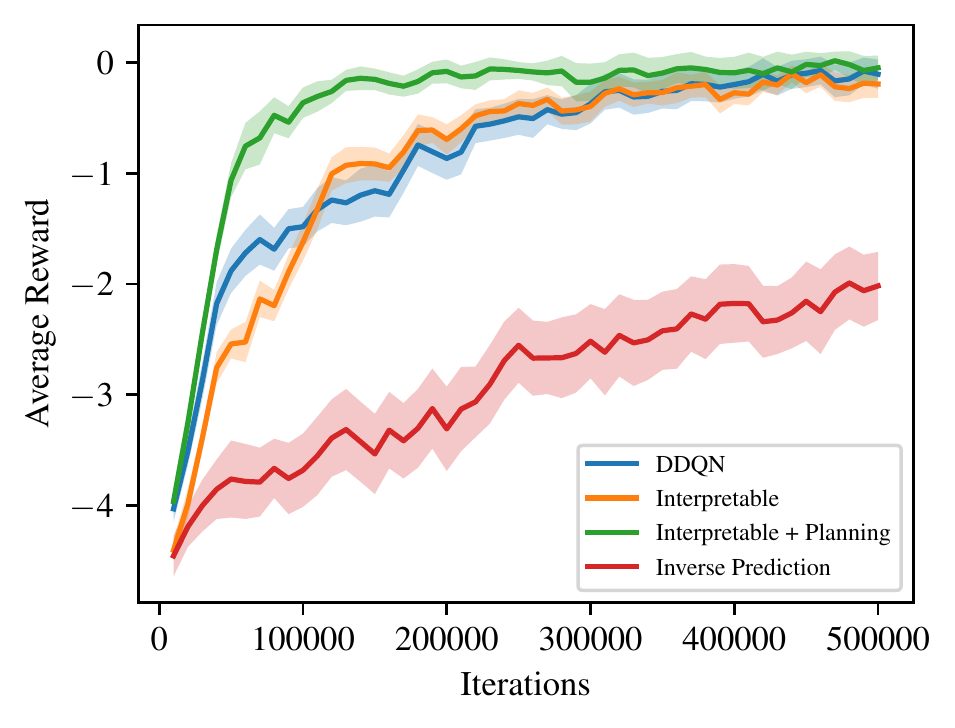}
\centering

     \caption{Performance of different (pre)trained representations on the random maze environment, measured as a mean (full line) and standard error (shaded area) over 5 seeds. The `Interpretable' setting uses an encoder pre-trained with 50k iterations to acquire a representation as in Fig.~\ref{fig:final_reps}, after which the encoder is frozen and a Q-network is trained on top with DDQN for 500k iterations. The `Interpretable + Planning' curve is similar to the `Interpretable' setting but uses DDQN with a planning algorithm in the controllable partition of the latent space with a depth of 3. The `DDQN' setting uses an encoder trained end-to-end with only DDQN for 500k iterations and the `Inverse Prediction' setting is equal to the 'Interpretable' setting but has an encoder pre-trained with $\loss_{inv}$ instead of $\loss_{\cont}$.}
     \label{fig:reward}
\end{figure}

\paragraph{Reinforcement Learning}
In order to verify whether a human-interpretable disentangled latent encoding is informative enough for downstream tasks, we formalize the random maze environment into an MDP with rewards. The agent acquires a reward $r_{t}$ of -0.1 at every time step, except when it finds the key in the top right part in which case it acquires a positive reward of 1. The episode ends whenever a positive reward is obtained or a total of 50 environment steps have been taken. For each new episode, a random wall structure is generated, and the agent starts over in the bottom left section of the maze (see Fig.~\ref{fig:reward}).
To see whether an interpretable disentangled latent representation is useful for RL, we compare different scenarios of (pre)training; (i) An encoder pretrained for 50k iterations to attain the representation in Fig.~\ref{fig:final_reps} and subsequently trained with DDQN for 500k iterations (ii) an encoder identical to the aforementioned but trained with DDQN and a planning algorithm (iii) an encoder pretrained for 50k iterations with $\loss_{inv}$ instead of $\loss_{c}$ and subsequently trained with DDQN for 500k iterations (iv) an encoder purely trained with DDQN gradients for 500k iterations. The resulting performances are compared in Fig.~\ref{fig:reward}. We find that a disentangled structured representation is suitable for downstream tasks, as it achieves comparable performance to training an encoder end-to-end with DDQN for 500k iterations. Although performance is similar, Fig.~\ref{fig:DDQN_reps} shows that an encoder updated solely with the DDQN gradient can lose any form of interpretability. Moreover, we show in Fig.~\ref{fig:reward} that a representation trained with an inverse prediction loss instead of a state-action forward prediction loss leads to poor downstream performance in the random maze environment. 

\paragraph{Planning}
 As seen in Fig.~\ref{fig:final_reps}, after pre-training with the unsupervised losses, an interpretable disentangled representation with the corresponding agent transitions is obtained. Due to this disentanglement of the controllable and uncontrollable features, we can for instance employ prior knowledge that the uncontrollable features in the maze environment are static, and employ latent planning in the controllable latent space only (see Fig.~\ref{fig:planning}). The planning algorithm used is derived from \cite{Oh2017ValueNetwork}, and is used to successfully plan only in the controllable partition of the latent representation $\z^{\cont}$, while freezing the input for $\z^{\uncont}$ regardless of planning depth. More details on the planning algorithm can be found in Appendix~\ref{AppendixA4_planning}. It can be observed that even when planning with a relatively small depth of 3, we achieve better performance than the pre-trained representation with an $\epsilon$-greedy policy and than the purely DDQN-updated encoder. 

\section{Limitations}

While the work presented here provides a step towards a better understanding of disentangling controllable and uncontrollable features within an encoder architecture, there remain some limitations that we must acknowledge, and which can provide a basis for future research.

First, our method's effectiveness was predominantly demonstrated on environments with relatively simple underlying dynamics. In these environments, the disentanglement process was easier to achieve due to the limited complexity of internal dynamics present. As we begin to transfer our approach to more complex environments characterized by more extensive internal dynamics, there can arise two problems; The first being that the separation of controllable from uncontrollable features may not be as clear-cut in more complex MDPs, but can be more on a spectrum, complicating the fundamental differences between a state-only and a state-action forward predictor. The second being that interpretability will be harder to enforce when there are a large number of underlying factors of variation. As distinct seeds can give different orderings and signs of the neurons in the final layer of the encoder, identifying a factor of variation can become exponentially harder for more complex environments.

Lastly, while our work showed that an action-conditioned forward predictor could be preferred over an inverse predictor in some environments for isolating controllable features, it may not hold for all scenarios. The inherent properties of different environments might show a necessity of using different predictors. Consequently, there could very well be MDPs where our current approach might not provide the same level of disentanglement showed in the MDPs used in this paper.

Despite these limitations, we believe our work provides a strong foundation upon which future research can build and further extend the possibilities of achieving a highly interpretable latent representation through disentanglement of controllable and uncontrollable features.

\section{Conclusion and Future Work}
We have shown the possibility of disentangling controllable and uncontrollable features in an encoder architecture, strongly increasing the interpretability of the latent representation while also showing the potential use of this for downstream learning and planning, even in a single latent partition.
This disentanglement of controllable and uncontrollable features in the latent representation of high-dimensional MDPs was achieved by propagating an action-conditioned forward prediction loss and a state-only forward prediction loss through distinct sections of the latent representation. Additionally, a contrastive loss and an adversarial loss were used to respectively avoid collapse and further disentangle the latent representation. Furthermore, we showed that an action-conditioned forward predictor can, in some environments, be preferred as compared to an inverse predictor in terms of isolating controllable features in the representation. Finally, by employing forward prediction in latent space, we were able to successfully run a planning algorithm while leveraging the properties of the environment. In particular, the disentanglement of controllable and uncontrollable features allowed us to keep $\z^{\uncont}$ frozen regardless of planning depth in the context of a distribution of randomly generated mazes, i.e. we only do forward prediction in $\z^{\cont}$.

\indent Future work could focus on gradually transferring our notion of disentanglement and interpretability to environments with more extensive underlying internal dynamics. Further work could also look at the ordering of the latent dimensions, as a latent representation is often arbitrarily ordered. This means that distinct seeds will lead to a different ordering and sign of the neurons in the final layer of the encoder. For example, if seed one would give agent position +x and +y for neurons 1 and 2 respectively, then seed two could give agent position -y and +x to the same neurons. As we are additionally using a contrastive loss while learning our representation, these results are compliant with the theory that a contrastive loss can recover the original latent information up to an orthogonal linear transformation \cite{contrastive_inverts}.

\begin{figure} [!t]
\vspace{-7mm}
     \centering
     \begin{subfigure}[b]{0.45\textwidth}
         \centering
         \includegraphics[width=\textwidth]{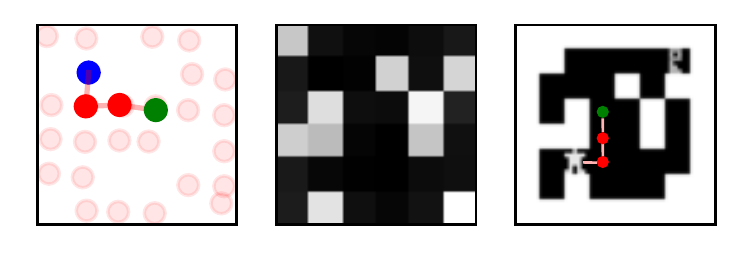}
         \caption{Planning depth 3}
     \end{subfigure}
     \hfill
     \begin{subfigure}[b]{0.45\textwidth}
         \centering
         \includegraphics[width=\textwidth]{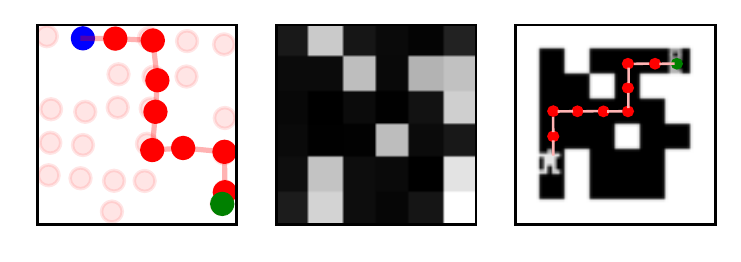}
         \caption{Planning depth 9}
     \end{subfigure}
     \caption{Visualization of the latent representation through an actual planning iteration utilizing a planning depth of 3 (a) and a planning depth of 9 (b), with the controllable representation $\z^{\cont} \in \mathbb{R}^{2}$ (left), the uncontrollable representation $\z^{\uncont} \in \mathbb{R}^{6\times6}$ (middle) that is kept static throughout planning depth and the original pixel input $s_{t} \in \mathbb{R}^{48\times 48}$ (right). The translucent red dots represent every possible encoded state in the random maze environment, the full blue dot represents the current encoded state, the red dots represent intermediate encoded states estimated by planning and the green dot represents the final predicted state as chosen by the planning algorithm, consistent with its depth.}
      \label{fig:planning}
\end{figure}

Certain benefits can be obtained as well with a particular design of the encoder architecture, as we have done in this paper using estimates of the necessary dimensions of $\z^{\cont}$ and $\z^{\uncont}$ for the different MDP environments. This can be seen as an inductive bias to aid disentanglement, as mentioned by \cite{scholkopf_locatello}. Succeeding work could also focus on finding more algorithmic benefits of this disentanglement of controllable/uncontrollable features in more complex environments.
For example, in the context of safety, a disentangled interpretable representation could allow incorporating latent state constraints in a planning algorithm. Lastly, as discussed by \cite{Glanois_Survey,scholkopf_locatello}, an interesting venue could be to further investigate the trade-off between interpretability and downstream performance. This is due to the fact that black-box representations such as Figure~\ref{fig:DDQN_reps} still seem to have excellent downstream performance with DDQN, where for the task of maze navigation, a human would perform substantially better using the representation portrayed in Figure~\ref{fig:final_reps} as compared to using the representation in Figure~\ref{fig:DDQN_reps}.

\bibliographystyle{IEEEtran}
\bibliography{root}

\vspace{24mm}
\appendix
\section{Additional Material} \label{Appendix}

\subsection{Ablation of the contrastive scalar}  \label{AppendixA1_entropyscalar}
Without using a pixel reconstruction loss, the contrastive loss $\loss_{H}$ is crucial in avoiding the trivial solution for any latent forward predictor \cite{Francois-Lavet2019CombinedRepresentations, Gelada2019DeepMDP:Learning}. The contrastive scalar that regulates the $\loss_{H}$ however remains the most influential hyperparameter. When $C_{d}$ is chosen too high, the representation remains in a compact cluster. On the other hand, when $C_{d}$ is chosen too low, unnecessary inter-sample distances are formed to enforce large individual latent distances. Two ablations of the contrastive scalar $C_{d}$ are shown in Fig.~\ref{fig:entropy_ablation}. 

\begin{figure} [h!]
     \centering
     \begin{subfigure}[b]{0.40\textwidth}
         \centering
         \includegraphics[width=\textwidth]{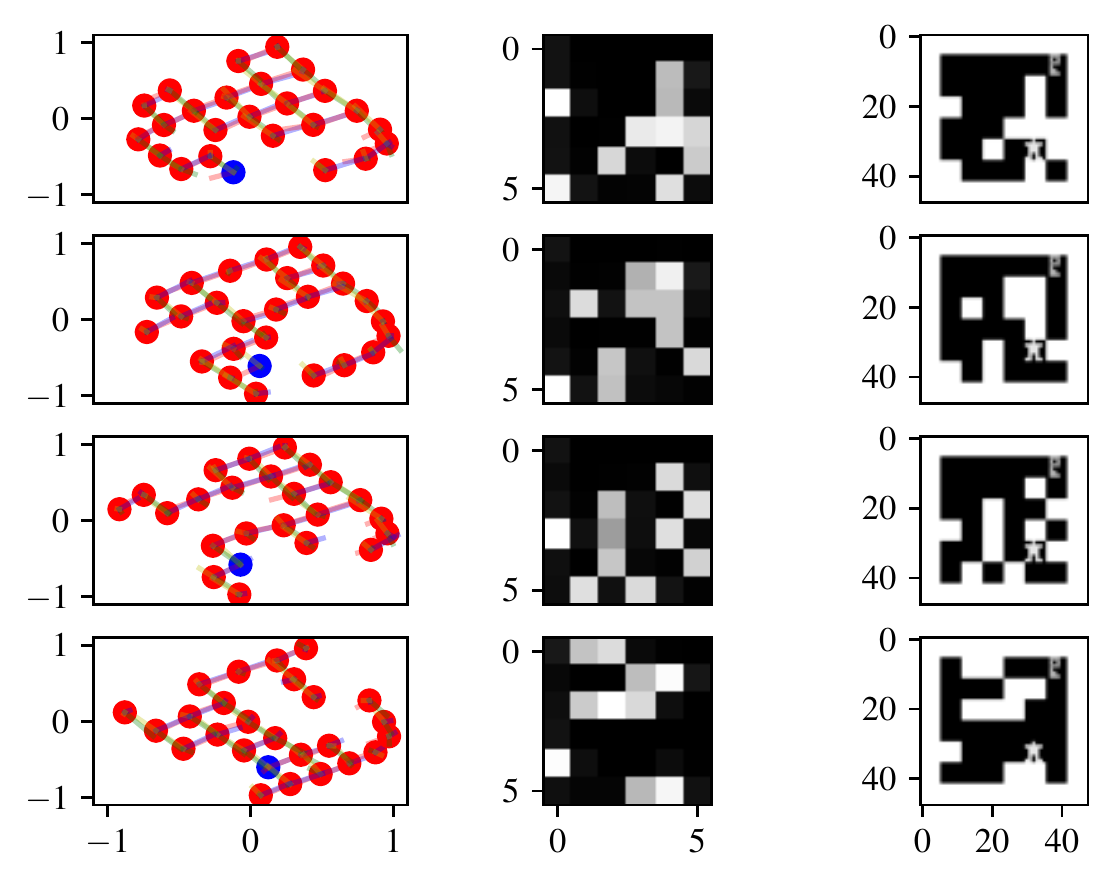}
         \caption{$C_{d}= 13$}
     \end{subfigure}
     \hfill
     \begin{subfigure}[b]{0.4\textwidth}
         \centering
         \includegraphics[width=\textwidth]{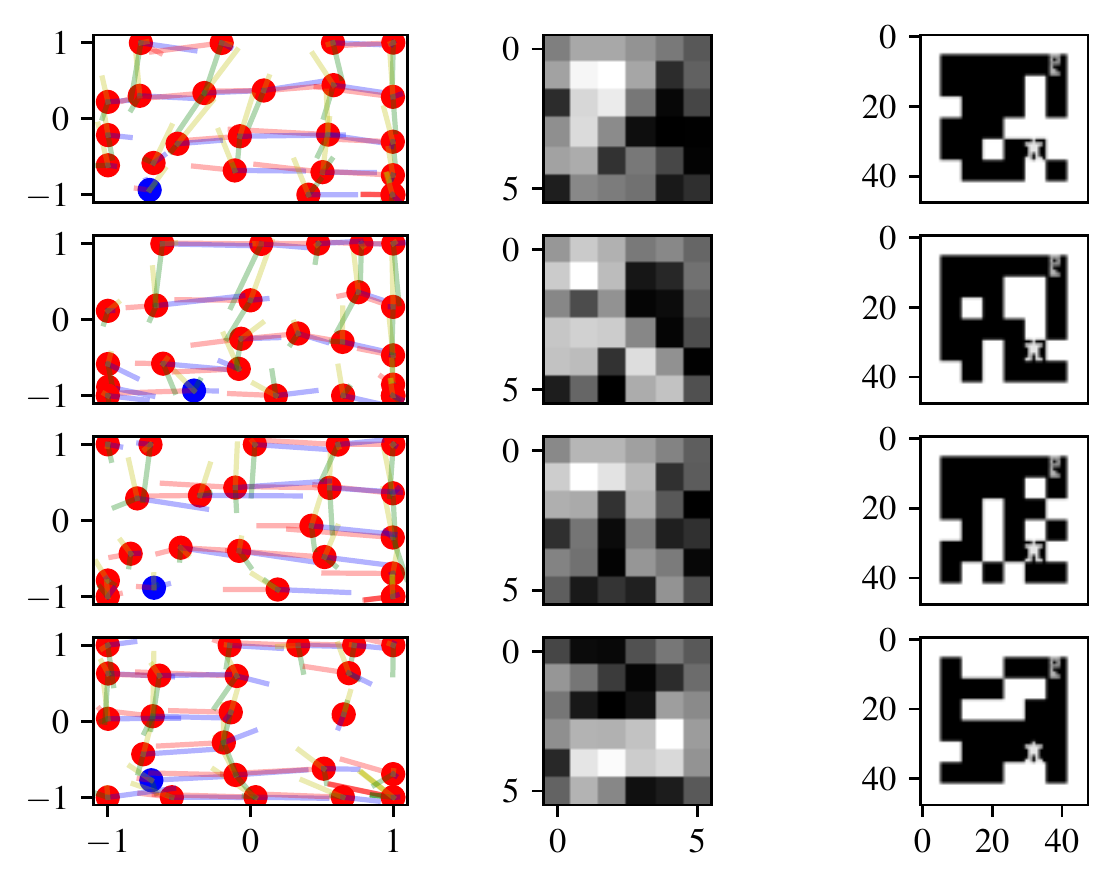}
         \caption{$C_{d} = 3$}
     \end{subfigure}
     \caption{Ablation of the hyperparameter $C_{d}$, where a higher value of $C_{d}$ enforces less entropy in the representation, while a lower value of $C_{d}$ especially pushes the controllable features $\z^{\cont}$ towards shapes that ensure large distances between samples. In both figures, the left column is $\z^{\cont} \in \mathbb{R}^{2}$, the middle column is $\z^{\uncont} \in \mathbb{R}^{6 \times 6}$ and the right column is the input state $\in \mathbb{R}^{48 \times 48}$.}
      \label{fig:entropy_ablation}
\end{figure}

\subsection{Ablation of learning rates}  \label{AppendixA2_learningrates}

We show experiments in Fig.~\ref{fig:lr_enc_ablation} and Fig.~\ref{fig:lr_sa_ablation} where we employ different learning rates for the encoder and the action-conditioned forward predictor, respectively.

\subsection{Ablation of the detachment of $\z^{\uncont}$ and ablation of the residual prediction}  \label{AppendixA3_detachment_residual}

As seen in the main paper in Figure~2, we detach the uncontrollable representation $\z^{\cont}$ from $\loss_{c}$ as we do not want controllable features to be present in $\z^{\uncont}$. We can see in Figure~\ref{fig:detachment} that updating $\z^{\uncont}$ with $\loss_{c}$ leads to slightly better transition predictions in $\z^{\cont}$, but also results in a less interpretable encoding of $\z^{\uncont}$. Furthermore, we can also see in Figure~\ref{fig:detachment} that, when using normal forward predictions instead of residual forward predictions, we lose almost all of our interpretable structure in $\z^{\uncont}$.

\begin{figure} [h!]
     \centering
     \begin{subfigure}[b]{0.4\textwidth}
         \centering
         \includegraphics[width=\textwidth]{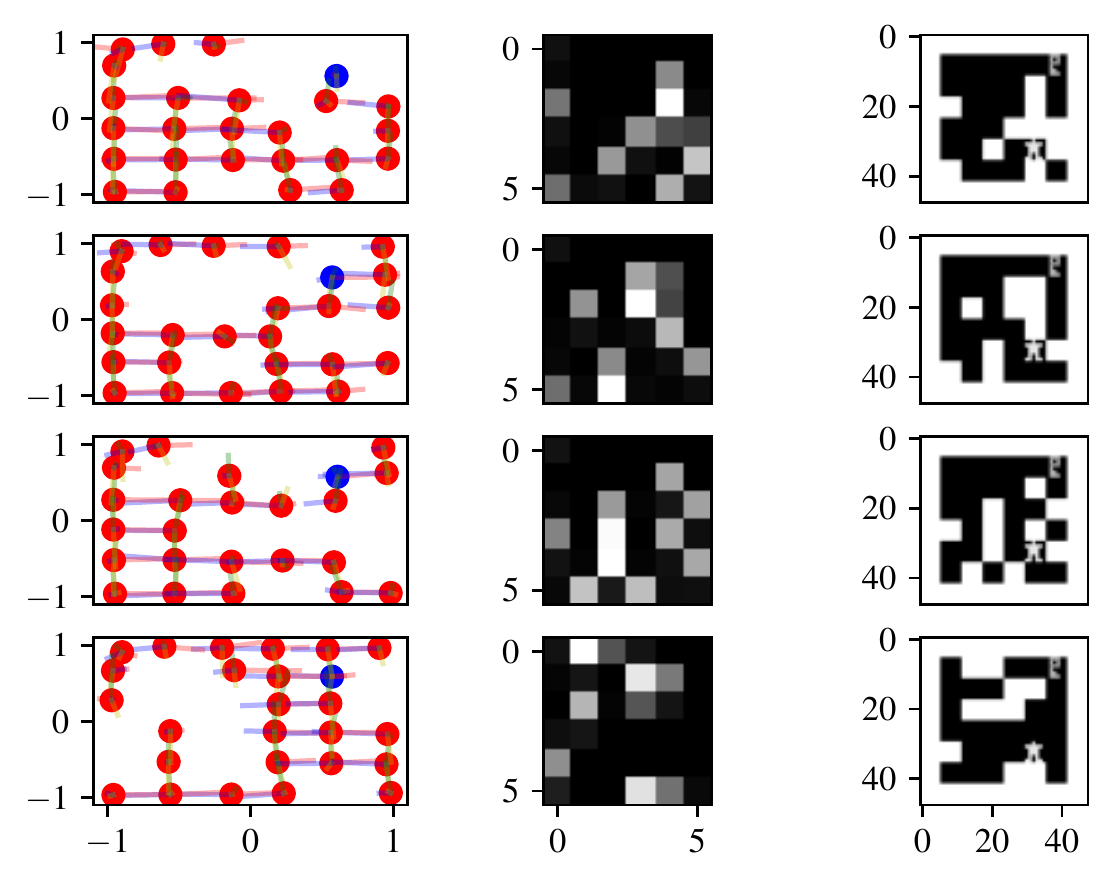}
         \caption{Encoder learning rate of 2e-4}
     \end{subfigure}
     \hfill
     \begin{subfigure}[b]{0.4\textwidth}
         \centering
         \includegraphics[width=\textwidth]{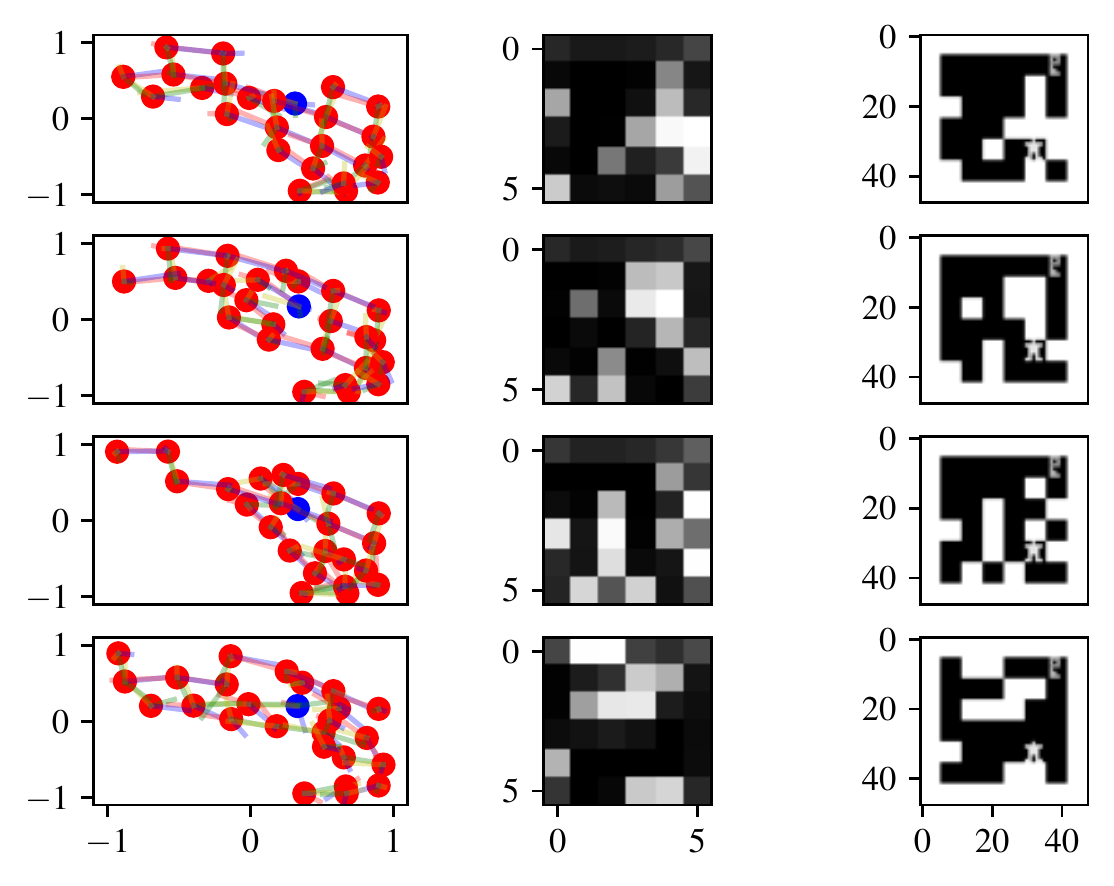}
         \caption{Encoder learning rate of 2e-6}
     \end{subfigure}
     \caption{Ablation of the learning rates for the encoder, where a too low learning rate causes collapse of $\z^{\cont}$ and a too high learning rate causes distortions in the uncontrollable features $\z^{\uncont}$. In both figures, the left column is $\z^{\cont} \in \mathbb{R}^{2}$, the middle column is $\z^{\uncont} \in \mathbb{R}^{6 \times 6}$ and the right column is the input state $\in \mathbb{R}^{48 \times 48}$.}
      \label{fig:lr_enc_ablation}
\end{figure}

\begin{figure} [h!]
     \centering
     \begin{subfigure}[b]{0.4\textwidth}
         \centering
         \includegraphics[width=\textwidth]{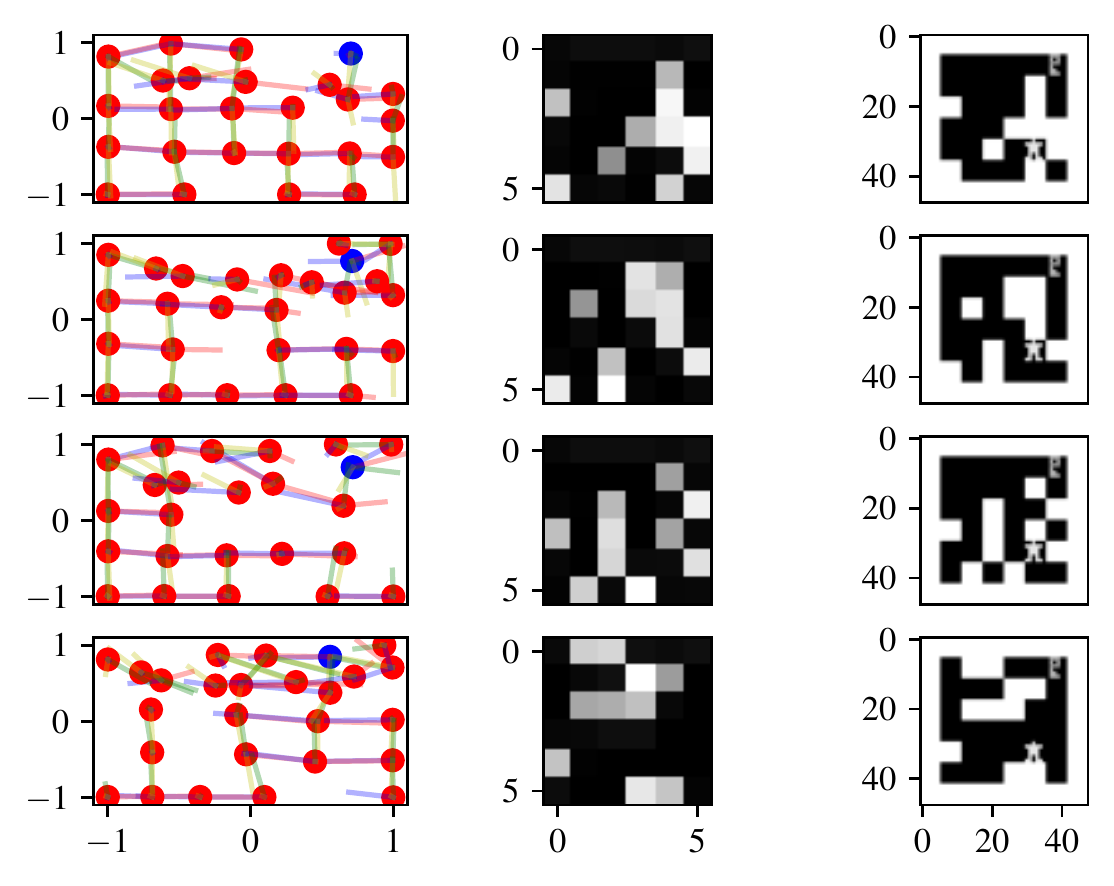}
         \caption{$\tr_{c}$ learning rate of 4e-4}
     \end{subfigure}
     \hfill
     \begin{subfigure}[b]{0.4\textwidth}
         \centering
         \includegraphics[width=\textwidth]{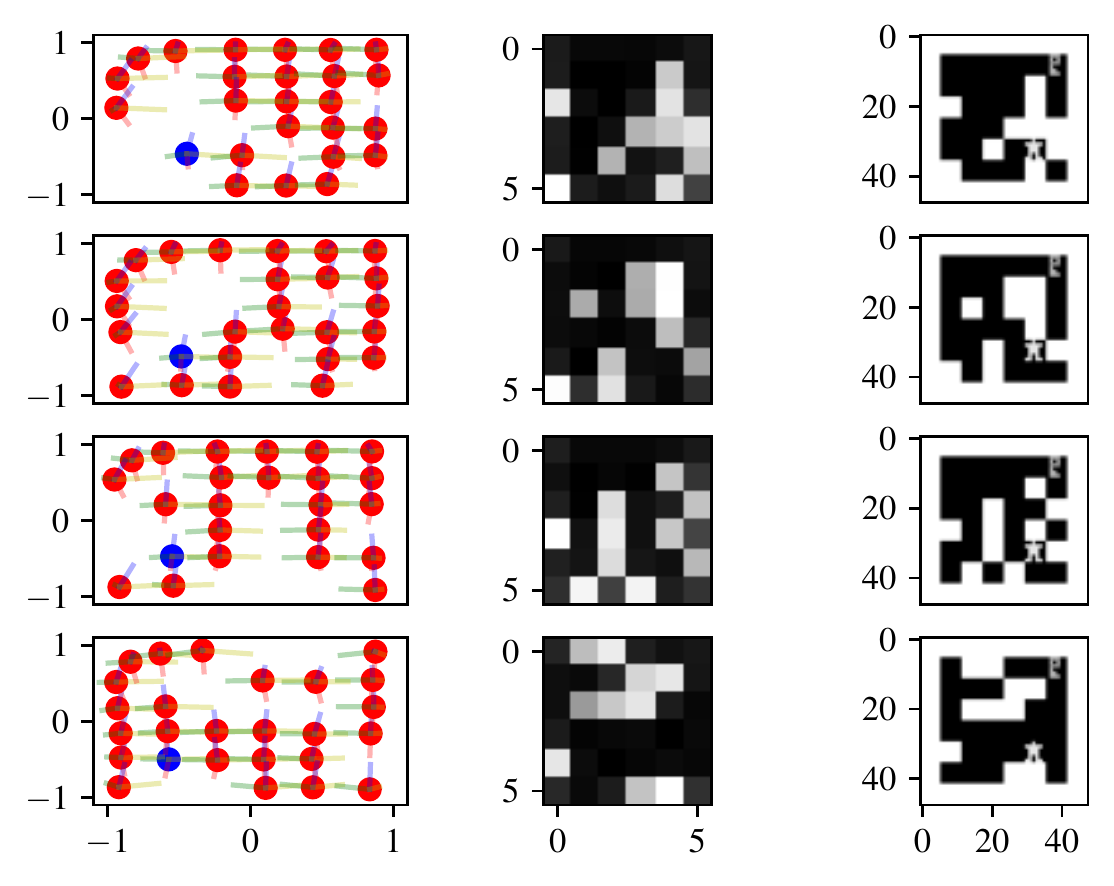}
         \caption{$\tr_{c}$ learning rate of 4e-6}
     \end{subfigure}
     \caption{Ablation of the learning rates for the action-conditioned forward predictor. A too high learning rate will cause the controllable representation to lose structure, while a low learning rate retains structure but does not learn strong transition dynamics. In both figures, the left column is $\z^{\cont} \in \mathbb{R}^{2}$, the middle column is $\z^{\uncont} \in \mathbb{R}^{6 \times 6}$ and the right column is the input state $\in \mathbb{R}^{48 \times 48}$.}
      \label{fig:lr_sa_ablation}
\end{figure}
\newpage

\begin{figure} [h!]
     \centering
     \begin{subfigure}[b]{0.4\textwidth}
         \centering
         \includegraphics[width=\textwidth]{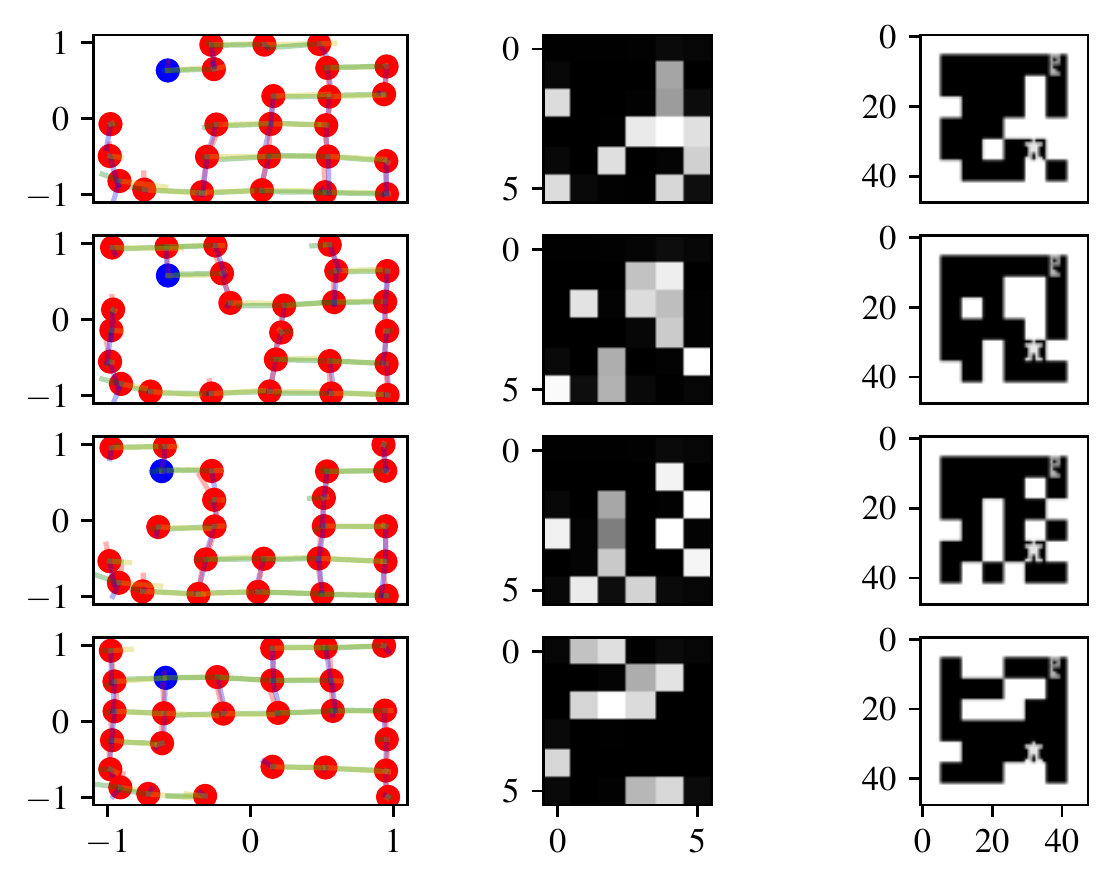}
         \caption{No detachment of $\z^{\uncont}$}
     \end{subfigure}
     \hfill
     \begin{subfigure}[b]{0.4\textwidth}
         \centering
         \includegraphics[width=\textwidth]{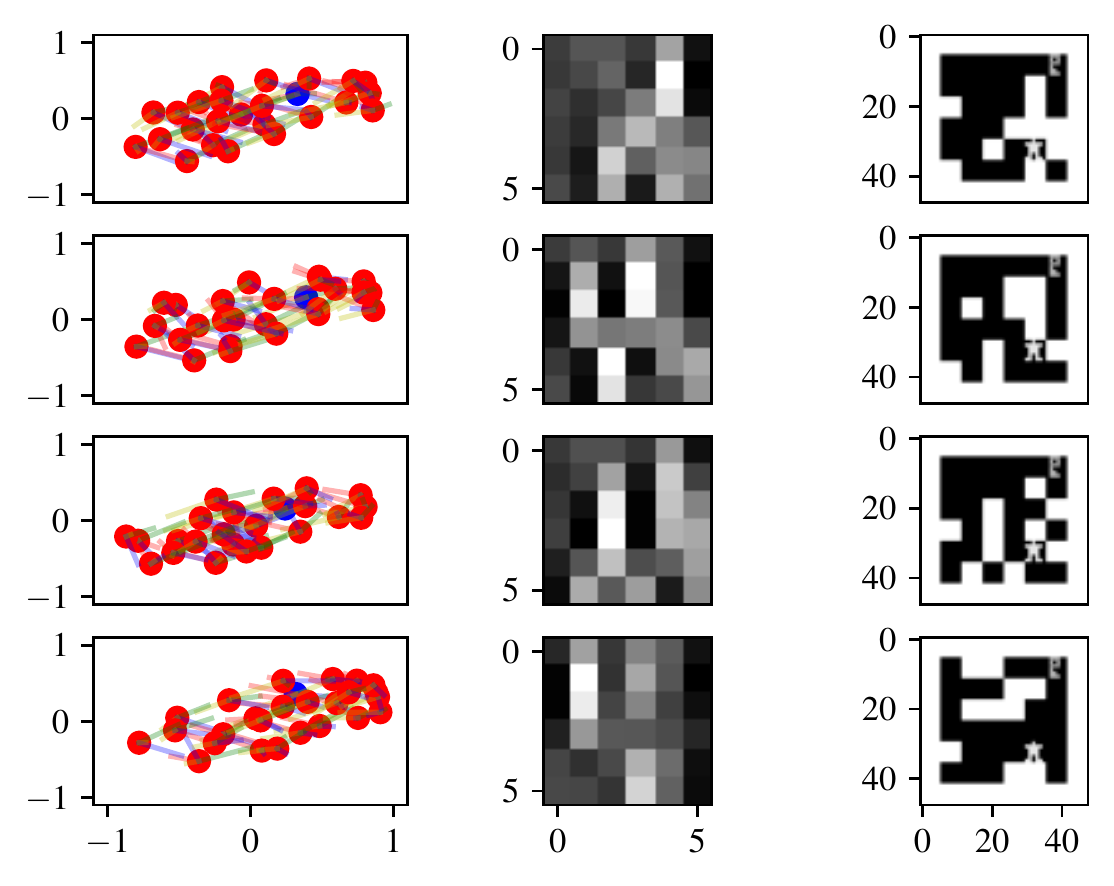}
         \caption{No residual predictions of $\z^{\cont}_{t+1}$ and $\z^{\uncont}_{t+1}$}
     \end{subfigure}
     \caption{In both figures, the left column is $\z^{\cont} \in \mathbb{R}^{2}$, the middle column is $\z^{\uncont} \in \mathbb{R}^{6 \times 6}$ and the right column is the input state $\in \mathbb{R}^{48 \times 48}$.}
      \label{fig:detachment}
\end{figure}

\subsection{Ablation of the entropy loss $\loss_{H2}$}  \label{AppendixA4_entropy_randommaze}

As the amount of possible encoded maze architectures goes to infinity due to the procedural generation, a collapse in the controllable features $\z^{\cont}$ can be noticed when using only $\loss_{H1}$ as the contrastive loss (see Fig.~\ref{fig:LH2}). On the other hand, when using only $\loss_{H2}$ as the contrastive loss, there is no more clear distinction in the uncontrollable representation $\z^{\uncont}$. The best results were obtained using a combination of the aforementioned losses.

\begin{figure} [h!]
     \centering
     \begin{subfigure}[b]{0.4\textwidth}
         \centering
         \includegraphics[width=\textwidth]{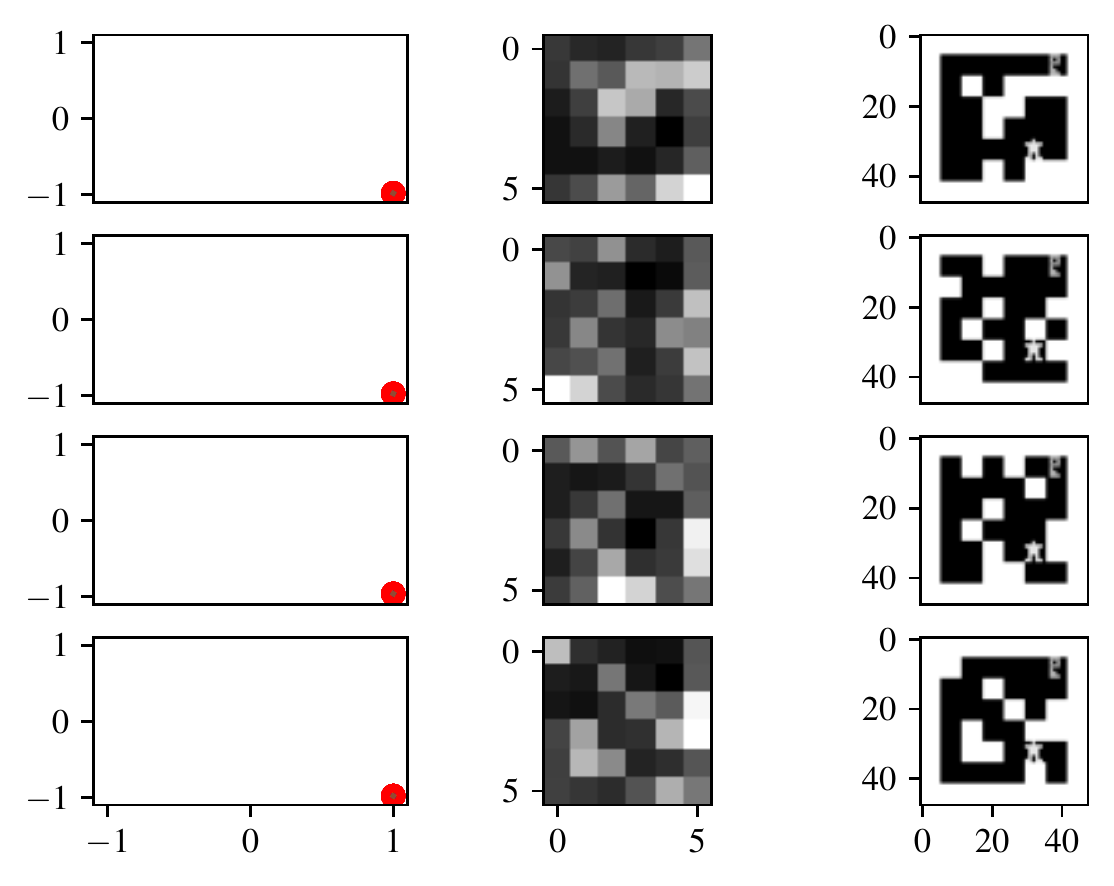}
         \caption{$\loss_{H} = \loss_{H1}$}
     \end{subfigure}
     \hfill
     \begin{subfigure}[b]{0.4\textwidth}
         \centering
         \includegraphics[width=\textwidth]{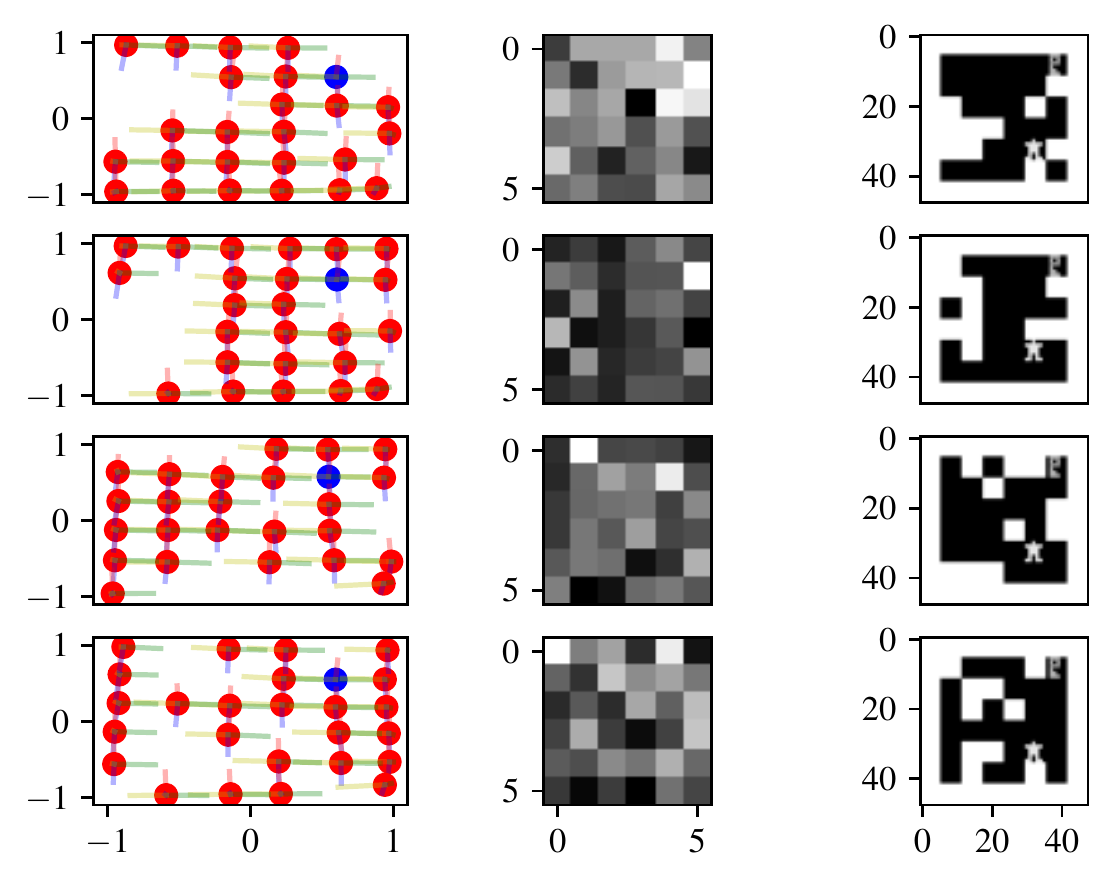}
         \caption{$\loss_{H} = \loss_{H2}$}
     \end{subfigure}
     \caption{In both figures, the left column is $\z^{\cont} \in \mathbb{R}^{2}$, the middle column is $\z^{\uncont} \in \mathbb{R}^{6 \times 6}$ and the right column is the input state $\in \mathbb{R}^{48 \times 48}$.}
      \label{fig:LH2}
\end{figure}


\subsection{Planning}  \label{AppendixA4_planning}
We use a planning algorithm derived from \cite{Oh2017ValueNetwork, Francois-Lavet2019CombinedRepresentations}, where we employ d-step planning as:


\begin{equation}
\fontsize{8pt}{11pt}\selectfont
    \hat Q^d((\hat{\z}^{\cont}_{t}, \z^{\uncont}), a)=\left\{
                \begin{array}{ll}
                  P((\hat{\z}^{\cont}_{t}, \z^{\uncont}),a;\theta_{r})+ \Gamma((\hat{\z}^{\cont}_{t}, \z^{\uncont}),a; \theta_{\gamma}) \ \underset{a' \in \mathcal A^*}{\operatorname{max}} \ \hat Q^{d-1}(    \\ (\hat{\z}^{\cont}_{t+1}, \z^{\uncont}),a'),\hspace{20mm}  \text{ if } d > 0\\
                  Q((\hat{\z}^{\cont}_{t}, \z^{\uncont}), a; \theta ), \hspace{18mm} \text{ if } d = 0
                \end{array}
              \right.
\label{eq:Qd}
\end{equation}

\begin{equation}
Q_{plan}^D((\hat{\z}^{\cont}_{t}, \z^{\uncont}), a)=\sum_{d=0}^{D}  \hat Q^d((\hat{\z}^{\cont}_{t}, \z^{\uncont}), a)
\label{eq:Qd2}
\end{equation}

Where $P(\obs_{t}, a;\theta_{r}): \mathcal{Z} \times \mathcal{A} \rightarrow \mathcal{R}$ represents the reward predictor and $\Gamma(\obs,a; \theta_{\gamma}): \mathcal{Z} \times \mathcal{A} \rightarrow \gamma$ represents the discount value predictor. The action is chosen by taking the argmax of $Q_{plan}^D((\hat{\z}^{\cont}_{t}, \z^{\uncont}), a)$. Note in the results from Section 5.3, we are only forward predicting in the controllable latent space $\z^{\cont}$, and that $\z^{\uncont}$ remains a fixed value regardless of planning depth. This is possible by making use of the prior knowledge of the maze environments together with a disentangled controllable and uncontrollable latent representation.


\subsection{Inverse Prediction}\label{appendix:Inverse}

 A common single-step inverse prediction is defined as:

\begin{equation}
    \hat{a}_{t} = f(\obs_{t}, \obs_{t+1})
\end{equation} \label{eq:inverse1}

where $\hat{a}_{t}$ is the predicted action and $f(\obs_{t}, \obs_{t+1})$ represents an arbitrarily structured function. In the random maze environment, we use a parameterized inverse predictor which predicts in latent space:

\begin{equation}
    \hat{a}_{t} = \inv(\z^{\cont}_{t}, \z^{\cont}_{t+1}, \z^{\uncont}_{t}, \z^{\uncont}_{t+1}; \theta_{inv})
\end{equation} \label{eq:inverse2}

Where $\inv(\cdot; \theta_{inv}) \in \mathcal{I}: \mathcal{Z} \rightarrow 
\mathcal{A}$ is a parameterized inverse prediction function. Since we have 4 actions, we use the 4-dimensional logit output $\hat{a}_{t}$ to calculate the inverse prediction loss $\loss_{inv}$ as:

\begin{equation} \label{eq:CELoss}
    S(\hat{a}_{i}) = \frac{\text{exp}({\hat{a}_{i}})}{\sum_{j=1}^{n_{a}} \text{exp}({\hat{a}_{j}})},
    \quad
    \loss_{inv} = - \sum_{i=1}^{n_{a}} a_{i}\log(S(\hat{a}_{i}))
\end{equation}

Here, $n_{a}$ is the number of actions, $S(\hat{a}_{i})$ represents the softmax operator and $a_{i}$ is the actual action, given as a 0 or 1 truth label. This is more commonly known as the Cross-Entropy loss computation.

\subsection{Reconstruction}  \label{AppendixA6_recon}

We run an additional ablation on the four mazes environment, where the contrastive loss $\loss_{H}$ is replaced with a pixel reconstruction loss. The resulting representation comparison can be seen in Fig.~\ref{fig:reconstruction}.

\begin{figure} [h!]
\vspace{-4mm}
     \centering
     \begin{subfigure}[b]{0.45\textwidth}
         \centering
         \includegraphics[width=\textwidth]{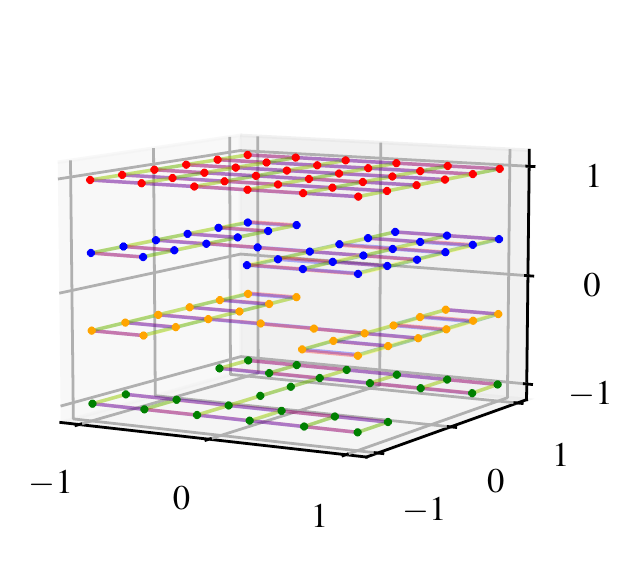}
         \caption{Contrastive loss}
     \end{subfigure}
     \hfill
     \begin{subfigure}[b]{0.45\textwidth}
         \centering
         \includegraphics[width=\textwidth]{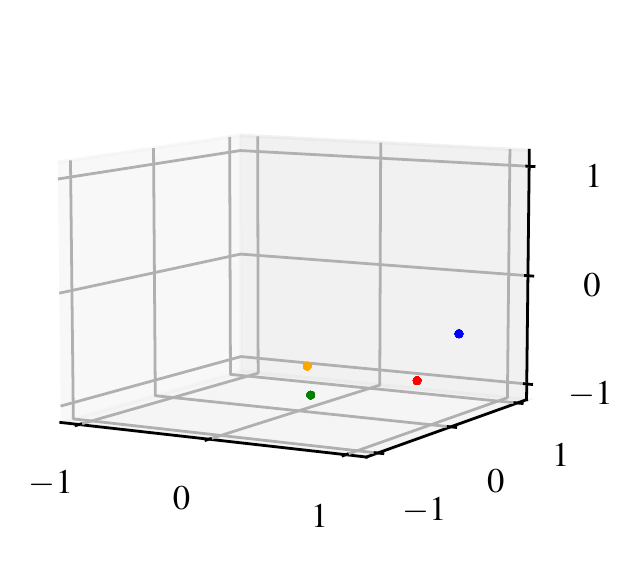}
         \caption{Reconstruction loss}
     \end{subfigure}
\caption{Visualization in a maze environment of the disentanglement of the controllable latent $\z^{\cont} \in \mathbb{R}^{2}$ on the horizontal axes, and the uncontrollable latent $\z^{\uncont} \in \mathbb{R}^{1}$ on the vertical axis, given for all states in the four maze environments shown in four different colors. The representation is trained on high-dimensional tuples $(\obs_{t}, a_{t}, r_{t}, \obs_{t+1})$, sampled from a replay buffer $\mathcal{B}$, gathered from random trajectories in the four maze environments. All possible states are encoded with $\z_{t} = \enc(\obs_{t};\theta_{enc}$) and plotted in (a) and (b) together with the transition prediction for each possible action. In (a), a clear disentanglement between the controllable agent's position and the uncontrollable wall architecture is portrayed. In (b), it seems that a reconstruction loss groups observations with similar pixel inputs together, and thus allows the forward predictors to 'collapse' to unit matrices, decreasing representation quality.}
      \label{fig:reconstruction}
\end{figure}

\newpage
\subsection{T-SNE}  \label{AppendixA7_tsne}

We conduct an additional experiment in the random maze environment where we use a latent dimension of 32, partition it in half to form $\z^{\cont} \in \mathbb{R}^{16}$ and $\z^{\uncont} \in \mathbb{R}^{16}$ and show the a T-SNE visualization of 6 different trajectories in random mazes in Fig.~\ref{fig:TSNE}. Note that, because the trajectories are random, only a subpart of the possible agent positions in every random maze is present.

\begin{figure} [h!]
     \centering
     \begin{subfigure}[b]{0.45\textwidth}
         \centering
         \includegraphics[width=\textwidth]{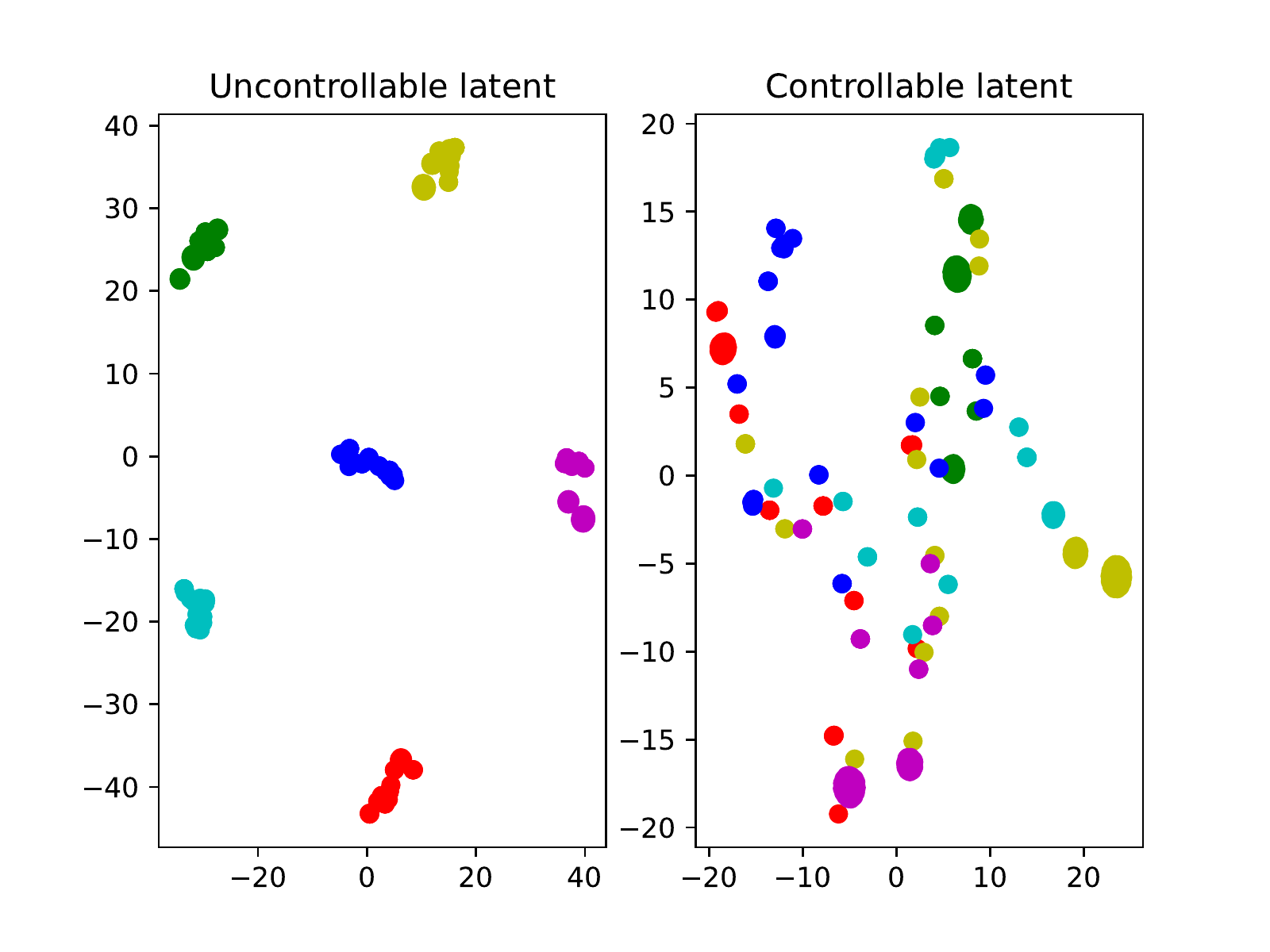}
        \caption{T-SNE of $\z^{\uncont}$ and $\z^{\cont}$}
     \end{subfigure}
     \begin{subfigure}[b]{0.45\textwidth}
     \vspace{2mm}
         \centering
            \includegraphics[width= 1.41in]{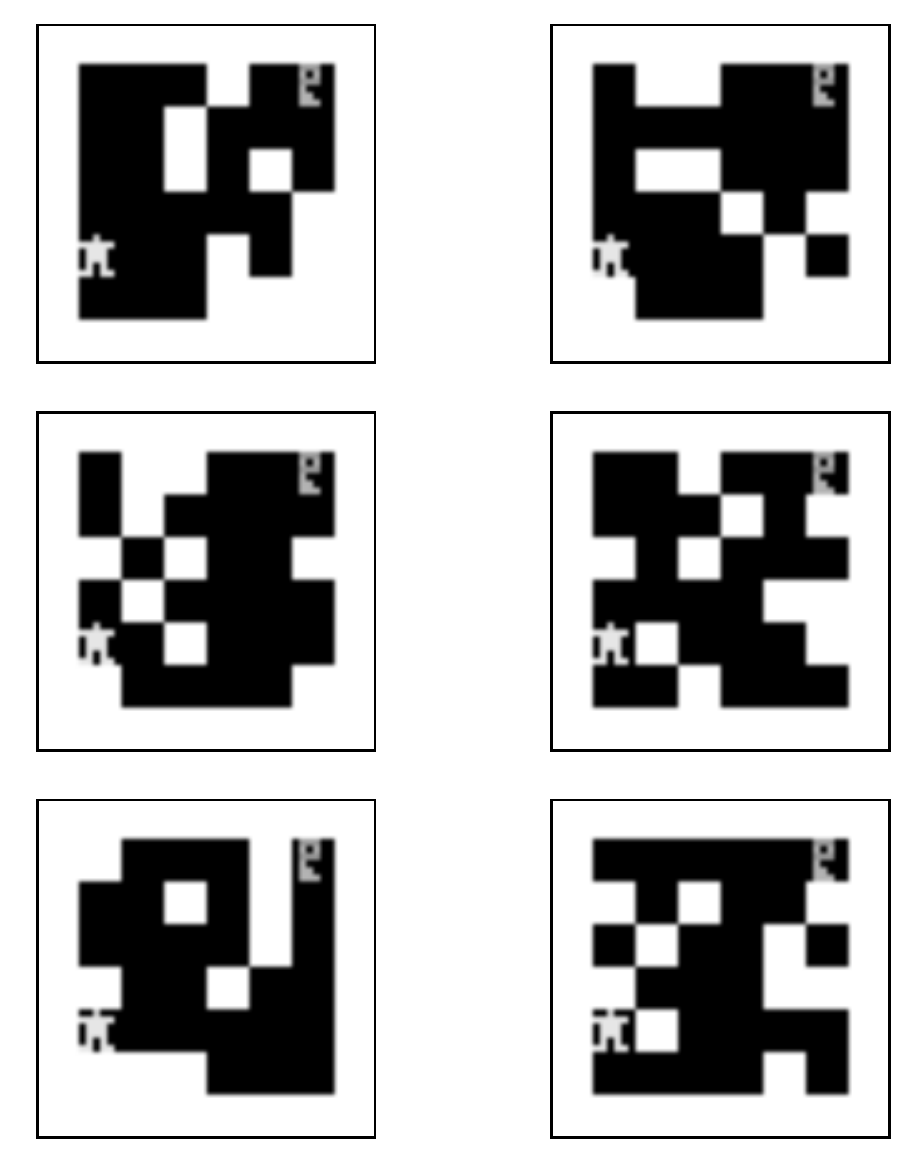}
            \vspace{5mm}
         \caption{6 random mazes}
     \end{subfigure}
     \caption{Ablation of a dimensionality increase in our random maze environment. Here, the total latent space is a 32-dimensional MLP output, where $\z^{\cont}$ and $\z^{\uncont}$ are both 16-dimensional. in (a), 6 random trajectories are plotted using T-SNE (perplexity=20) in different colors for both $\z^{\cont}$ and $\z^{\uncont}$, where  $\z^{\uncont}$ remains similar across a trajectory (same wall architecture), and $\z^{\cont}$ differs across the trajectory (different agent positions). In (b), a collection of the random mazes are shown from which the random trajectories have been taken. }
     \label{fig:TSNE}
     \end{figure}

     

\section{Experiment details}  \label{Appendix:hyperparams}

The Pytorch framework was used for all experiments, as well as the Adam optimizer \cite{Kingma2014Adam:Optimization}. We employ a batch size of 32 tuples $(\obs_{t}, a_{t}, r_{t}, \obs_{t+1})$ for every update. In all experiments, we detach $\z^{\cont}_{t}$ in the calculation of $\loss_{c}$, as it allowed us to use a larger learning rate for $\tr_{c}$ without causing instabilities.

\paragraph{Simple Maze}
The replay buffer $\mathcal{B}$ is filled with 5k transitions from each of the four wall architectures. The transitions are collected by the agent following a random policy. The learning rate for the encoder is $5\cdot 10^{-5}$, for the action-conditioned forward predictor $1\cdot 10^{-3}$ and for the uncontrollable forward predictor $5\cdot 10^{-5}$. The contrastive scalar $C_{d}$ is set to 15.

\paragraph{Catcher}
The replay buffer $\mathcal{B}$ is filled with 25k transitions. The transitions are collected by the agent following a random policy. A new random maze is created after 50 time steps or when the reward is acquired. The learning rate for the encoder is $2\cdot 10^{-5}$, for the action-conditioned forward predictor $4\cdot 10^{-5}$ and for the uncontrollable forward predictor $1\cdot 10^{-5}$. When using the adversarial loss, we use a learning rate of $1\cdot 10^{-3}$ for the adversarial predictor. The contrastive scalar $C_{d}$ is set to 5.

\paragraph{Random Maze}
The replay buffer $\mathcal{B}$ is filled with 50k transitions, representing around 1000 maze architectures. The transitions are collected by the agent following a random policy. The learning rates used are equal to those of the catcher environment; for the encoder $2\cdot 10^{-5}$, for the action-conditioned forward predictor $4\cdot 10^{-5}$ and for the uncontrollable forward predictor $1\cdot 10^{-5}$. After freezing the encoder, we train the action-conditioned forward predictor for an additional 250k iterations on the same 50k transitions in the buffer $\mathcal{B}$. For updating the Q-network with DDQN, we use a learning rate of $1\cdot 10^{-4}$, and a $\tau$ of 0.02. The contrastive scalar $C_{d}$ is set to 13. When using planning, we employ a learning rate of $5\cdot 10^{-5}$ for the reward and discount prediction networks.

\paragraph{Contrastive Loss} For the catcher and random maze environment, given that $\z^{\cont}$ is 1 or 2-dimensional, and $\z^{\uncont}$ is a 36-dimensional feature map, we alleviate dimensional mismatch when calculating the contrastive loss in Equation 4 in the main paper. This is done by taking a random subset of 15 out of 36 feature values in $\z^{\uncont}$ for every batch.
      
\section{Network Architecture} \label{app:architecture}
We use the same base encoder for all experiments, made up of 2 convolutional layers of 32 channels each, with a kernel size of 3 and stride 2, except for the final layer which has stride 1. Both convolutional layers have a Rectified Linear Unit (ReLU) nonlinear activation.

In the quadruple maze environment, the output of the base convolutional encoder is flattened and used as an input to a single linear layer with 3 outputs ($\z^{\cont} + \z^{\uncont}$) and a hyperbolic tangent  (tanh) activation function.

In the catcher and random maze environments, we use the following encoder head to extract the uncontrollable features; the base convolutional layers are followed by a single convolutional layer with 32 channels, a kernel size of 4 and a stride of 1. This layer is followed by a ReLU activation function and an AveragePool layer with an output size of 6. For the controllable features, we flatten the output of the base convolutional encoder and use this as an input to a linear layer with 200 neurons and a tanh activation function. This layer is followed by another linear layer with $n_{c}$ neurons and a tanh activation function.

The transition and prediction models all have the same structure, with linear layers of 32-128-128-32-x neurons where x is the output dimension in line with the predicted feature's dimension. The linear layers all have tanh activation functions except for the final output. Only the action-conditioned transition predictor of the random maze environment has larger layer sizes, with linear layers of 128-512-512-128-2, to account for slightly more complicated transitions. The DQN network used is of size 128-512-512-128-4, with an output value corresponding to each possible action.

\newpage

\subsection{Quadruple Maze Progression} \label{appendix:fourmaze}

\begin{figure} [h!]
     \centering
     \begin{subfigure}[b]{0.3\textwidth}
         \centering
         \includegraphics[width=\textwidth]{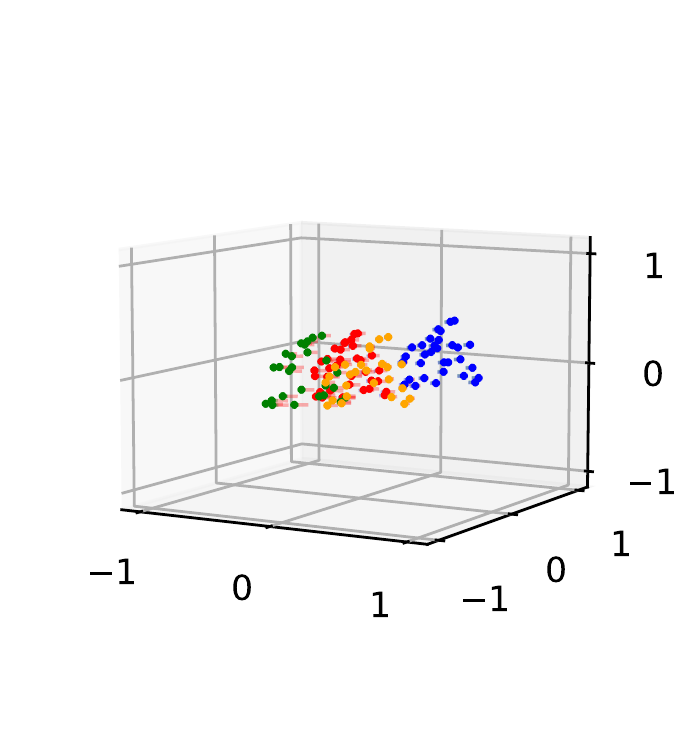}
         \caption{1k iterations}
     \end{subfigure}
     \begin{subfigure}[b]{0.3\textwidth}
         \centering
         \includegraphics[width=\textwidth]{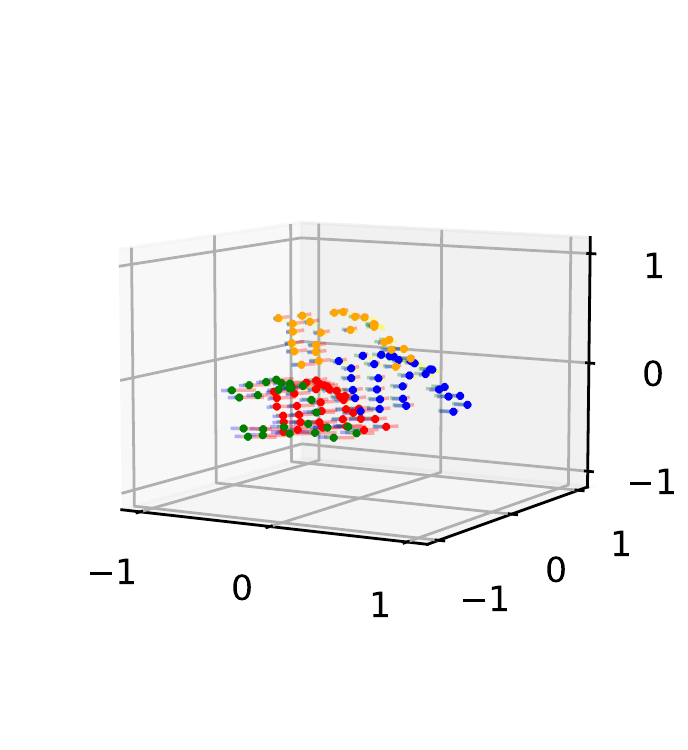}
         \caption{2k iterations}
     \end{subfigure}
          \begin{subfigure}[b]{0.3\textwidth}
         \centering
         \includegraphics[width=\textwidth]{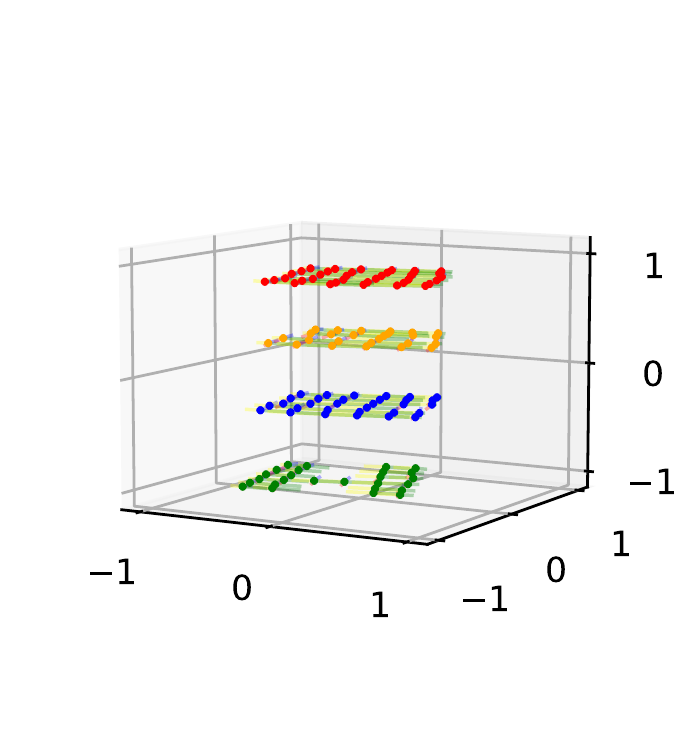}
         \caption{5k iterations}
     \end{subfigure}
     \caption{Progression of the separation of the controllable $\z^{\cont}$ (x and y-axis) and uncontrollable $\z^{\uncont}$ (z-axis) features in the maze environment.}
      \label{fig:fourmaze_progression}
\end{figure}

\newpage
\subsection{Catcher} 

\label{appendix:catcher}
\centering
\begin{figure}[h!]
  \begin{subfigure}[b]{0.4\textwidth}
    \subcaptionbox{$\z^{\cont} \in \mathbb{R}^{1}$ without $\loss_{adv}$}{
    \includegraphics[width=0.71\textwidth]{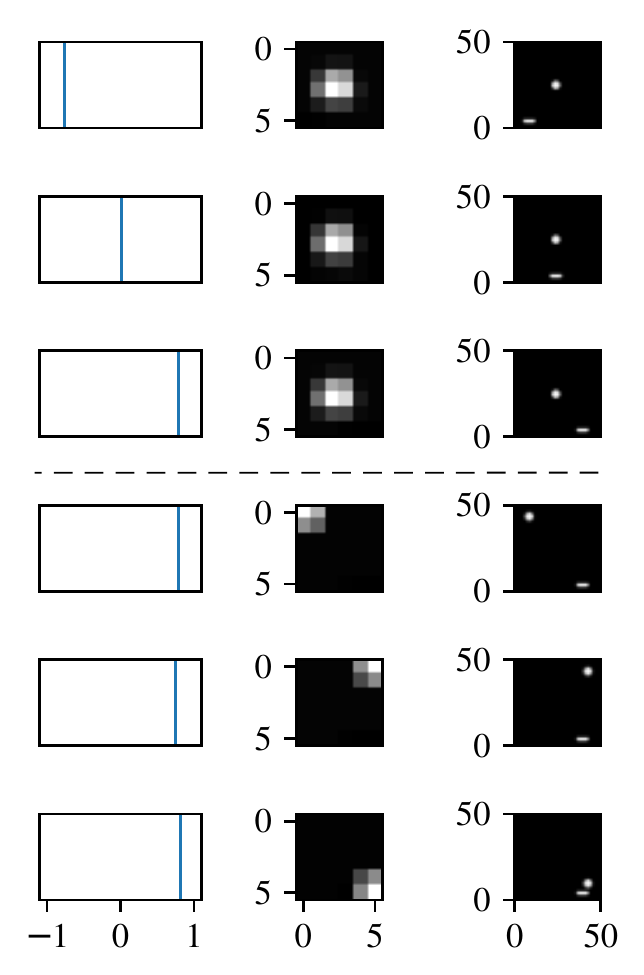}\hspace*{2em}%
    }
  \end{subfigure}
  \centering
  \begin{subfigure}[b]{0.4\textwidth}
        \subcaptionbox{$\z^{\cont} \in \mathbb{R}^{2}$ without $\loss_{adv}$}{
    \includegraphics[width=0.71\textwidth]{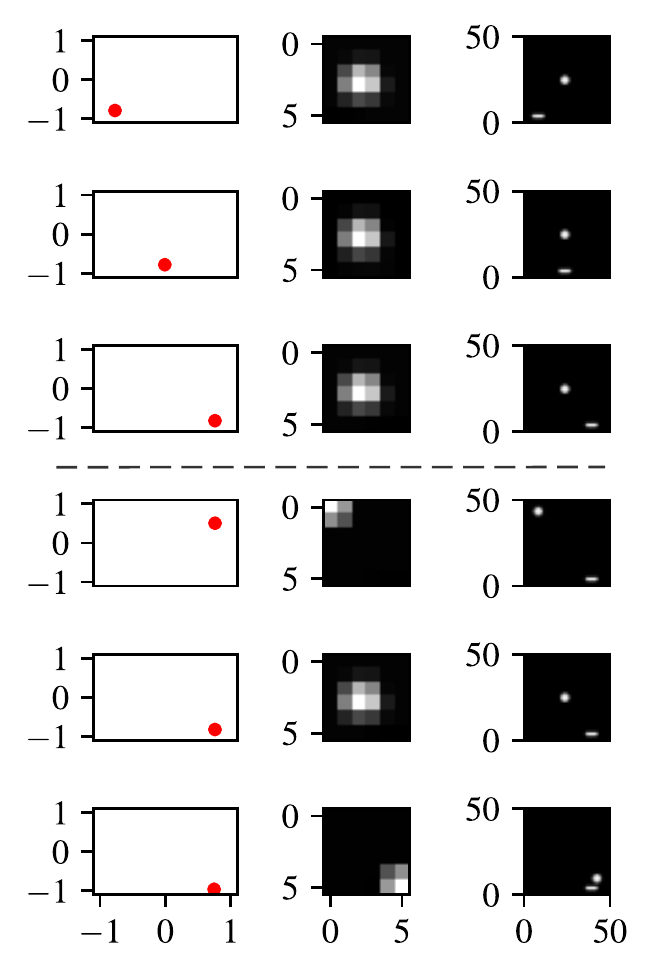}\hspace*{2em}%
    }
  \end{subfigure}
        \caption{Comparison of training the representation for the catcher environment with either 1 or 2-dimensions for the controllable representation $\z^{\cont}$. When using more dimensions for $\z^{\cont}$ than needed, it can be observed that some information of the ball position can be present in $\z^{\cont}$.}
      \label{fig:catcherablation}
      \end{figure}

\end{document}